\pdfoutput=1
\documentclass[11pt,a4paper]{article}
\usepackage[T1]{fontenc}
\usepackage{lmodern}
\usepackage[a4paper,margin=2.5cm]{geometry}
\usepackage{microtype}
\usepackage{graphicx}
\usepackage{amsmath,amssymb,amsfonts}
\usepackage{xcolor}
\usepackage{booktabs}
\usepackage{makecell}
\usepackage{acro}
\usepackage{fontawesome5}
\usepackage[numbers,sort&compress]{natbib}
\usepackage[font=small,labelfont=bf,labelsep=quad,singlelinecheck=false]{caption}
\definecolor{linkblue}{RGB}{0,70,150}
\usepackage[colorlinks,linkcolor=linkblue,citecolor=linkblue,urlcolor=linkblue]{hyperref}
\hypersetup{pdftitle={6 Fingers, 1 Kidney: Natural Adversarial Medical Images Reveal Critical Weaknesses of Vision-Language Models}, pdfauthor={Leon Mayer, Piotr Kalinowski, Caroline Ebersbach, Marcel Knopp, Tim Rädsch, Evangelia Christodoulou, Annika Reinke, Fiona R. Kolbinger, Lena Maier-Hein}}

\renewcommand{\figurename}{Fig.}
\newcommand{\bmhead}[1]{\paragraph*{#1}}
\newcommand{\backmatter}{}
\newcommand{\botrule}{\bottomrule}

\newcommand{\meanCIpar}[3]{#1\% {\tiny (#2\%, #3\%)}}
\newcommand{\meanCIstacked}[3]{\makecell{#1\% \\{\tiny (#2\%, #3\%)}}}
\newcommand{\bestCIpar}[3]{$\mathbf{#1}$\% {\tiny (#2\%, #3\%)}}
\newcommand{\almostbestCIpar}[3]{\underline{#1}\% {\tiny (#2\%, #3\%)}}
\DeclareAcronym{VLM}{ short = VLM, long = vision-language model }
\DeclareAcronym{AI}{ short = AI, long = artificial intelligence }
\DeclareAcronym{MRI}{ short = MRI, long = magnetic resonance imaging }
\DeclareAcronym{MRA}{ short = MRA, long = magnetic resonance angiography }
\DeclareAcronym{CT}{ short = CT, long = computed tomography }
\DeclareAcronym{MoE}{ short = MoE, long = mixture-of-experts }
\begin{document}
\begin{center}
{\LARGE\bfseries 6 Fingers, 1 Kidney: Natural Adversarial Medical Images Reveal Critical Weaknesses of Vision-Language Models\par}
\vspace{1.4em}
{\large Leon~Mayer$^{1,2,*,\dagger}$, Piotr~Kalinowski$^{1,3,4,\dagger}$, Caroline~Ebersbach$^{1,2}$, Marcel~Knopp$^{1,3}$, Tim~Rädsch$^{1,5,6,7}$, Evangelia~Christodoulou$^{1}$, Annika~Reinke$^{1,5}$, Fiona R.~Kolbinger$^{8,9}$, Lena~Maier-Hein$^{1,3,4,5,10,11,12,*}$\par}
\vspace{1em}
{\footnotesize $^{1}$German Cancer Research Center (DKFZ) Heidelberg, Division of Intelligent Medical Systems, Germany\\
$^{2}$Medical Faculty, Heidelberg University, Germany\\
$^{3}$Faculty of Mathematics and Computer Science, Heidelberg University, Germany\\
$^{4}$HIDSS4Health - Helmholtz Information and Data Science School for Health, Karlsruhe/Heidelberg, Germany\\
$^{5}$Helmholtz Imaging, German Cancer Research Center (DKFZ), Germany\\
$^{6}$Engineering Faculty, Heidelberg University, Germany\\
$^{7}$School of Computation, Information and Technology, TUM, Germany\\
$^{8}$Weldon School of Biomedical Engineering, Purdue University, West Lafayette, IN, USA\\
$^{9}$Department of Visceral, Thoracic and Vascular Surgery, University Hospital and Faculty of Medicine Carl Gustav Carus, TUD Dresden University of Technology, Dresden, Germany\\
$^{10}$National Center for Tumor Diseases (NCT), NCT Heidelberg, a partnership between DKFZ and University Hospital Heidelberg, Germany\\
$^{11}$Heidelberg University Hospital, Surgical Clinic, Surgical AI Research Group, Heidelberg, Germany\\
$^{12}$Mohamed Bin Zayed University of Artificial Intelligence (MBZUAI), Abu Dhabi, UAE\par}
\vspace{0.6em}
{\footnotesize $^{*}$Corresponding authors: \href{mailto:leon.mayer@dkfz-heidelberg.de}{leon.mayer@dkfz-heidelberg.de}, \href{mailto:l.maier-hein@dkfz-heidelberg.de}{l.maier-hein@dkfz-heidelberg.de}\\ $^{\dagger}$Equal contribution. Each co-first author may list themselves as lead author on their CV.\par}
\vspace{1.4em}
{\setlength{\fboxsep}{7pt}\fcolorbox{black!25}{black!4}{\begin{tabular}{@{}l@{\hspace{0.5em}}l@{\hspace{0.6em}}l@{}}
\raisebox{-0.1em}{\large\faGithub} & \textbf{Code} & \href{https://github.com/IMSY-DKFZ/adversarialanatomybench}{\texttt{github.com/IMSY-DKFZ/adversarialanatomybench}}\\[4pt]
\raisebox{-0.25em}{\includegraphics[height=1.15em]{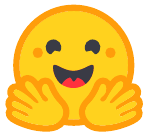}} & \textbf{Dataset} & \href{https://huggingface.co/datasets/IMSY-DKFZ/AdversarialAnatomyBench}{\texttt{huggingface.co/datasets/IMSY-DKFZ/AdversarialAnatomyBench}}
\end{tabular}}\par}
\end{center}
\vspace{0.4em}

\begin{abstract}
\Acp{VLM} are increasingly integrated into clinical workflows. However, existing benchmarks primarily assess performance on common anatomical presentations and fail to capture the challenges posed by rare variants. To address this gap, we introduce \textit{AdversarialAnatomyBench}, the first benchmark comprising naturally occurring rare anatomical variants across diverse imaging modalities and anatomical regions. We call such variants that violate learned priors about “typical” human anatomy natural adversarial anatomy.

Benchmarking 25 state-of-the-art \acsp{VLM} with \textit{AdversarialAnatomyBench} yielded three key insights. First, when queried with basic medical perception tasks, mean accuracy dropped from 71\% on typical to 28\% on atypical anatomy. Even the best-performing models, GPT-5, Gemini 2.5 Pro, and Llama 4 Maverick, showed performance drops of 39-51\%. Second, model errors closely mirrored expected anatomical biases. Third, neither model scaling nor interventions, including bias-aware prompting and test-time reasoning, resolved these issues.
These findings highlight a critical limitation in current \acsp{VLM}: their poor generalization to rare anatomical presentations. \textit{AdversarialAnatomyBench} provides a foundation for systematically measuring and mitigating anatomical bias in multimodal medical \acf{AI} systems.
\end{abstract}

\section{Introduction}\label{sec1}
\Acp{VLM} are increasingly deployed across clinical decision-support systems, diagnostic assistance platforms, and medical education, reflecting a broader shift toward multimodal \ac{AI} in healthcare. As \acp{VLM} move from research prototypes to clinical tools, ensuring their reliability across the full spectrum of anatomical presentations becomes paramount.

However, the spectrum of anatomical phenotypes includes anatomical variants that deviate from textbook anatomy. In this paper, we will refer to these variants as \textit{atypical}, while we will refer to common phenotypes as \textit{typical}. While atypical variants are individually rare, they collectively affect a substantial portion of patients. For example, polydactyly, the presence of supernumerary digits, occurs in approximately 1 in 800 to 2,700 births \cite{bubshait2022review}. Horseshoe kidney, where the two kidneys are fused at their lower poles, affects roughly 1 in 500 individuals \cite{taghavi2016horseshoe}. Situs inversus, a complete mirror-image reversal of organ positioning, occurs in approximately 1 in 10,000 people \cite{eitler2022situs}. These and other anatomical variants span diverse organ systems and manifest across clinical imaging modalities, thus affecting all domains of medical care.

The clinical stakes of misdiagnosing rare anatomical presentations are considerable, as failure to recognize such variants can lead to inappropriate treatment planning or delayed diagnosis of associated pathologies. In consequence, clinical \ac{AI} systems must demonstrate robust performance not only on the common anatomical presentations that dominate training data, but also on the atypical variants. This requirement extends beyond accuracy on typical cases to encompass reliable generalization across the full range of human anatomical diversity, which is pivotal to reduce bias and facilitate safe real-world deployment.

\acp{VLM} trained on internet-scale data learn strong statistical priors that favor typical anatomy, reflecting training distributions heavily skewed toward typical presentations such as five fingers, two separate kidneys, and standard organ positioning. Unlike traditional medical \ac{AI} systems designed for single tasks within narrow domains, these foundation models must generalize across diverse medical scenarios, creating a fundamental tension: learned priors that enhance performance on common cases can override critical visual evidence when encountering rare variants. Current evaluation frameworks do not systematically compare performance on typical and atypical presentations, obscuring a critical limitation in model reliability \cite{royer2024multimedeval,zhou2025drvd}.

This work has been inspired by recent progress in the general computer vision community. A recent line of work demonstrated that \acp{VLM} in general domains exhibit strong prior bias that overrides visual evidence \cite{vo2025vision}, and that in the medical domain, \acp{VLM} exploit textual cues over visual information and display shortcut behaviors \cite{buckley2023multimodal,gu2025illusion}. Another body of work within the medical \ac{AI} community has primarily focused on clinical attack scenarios \cite{sadanandan2025vsf}, demographic biases, and performance disparities across marginalized groups \cite{yang2025demographic, xu2025biasicl}. However, no systematic evaluation of \ac{VLM} performance on naturally occurring rare anatomical variants in medical imaging has been conducted to date. It remains unknown whether anatomical priors learned from typical anatomy cause systematic errors when models encounter atypical presentations.

\begin{figure}[!htbp]
\centering
\includegraphics[width=0.9\textwidth]{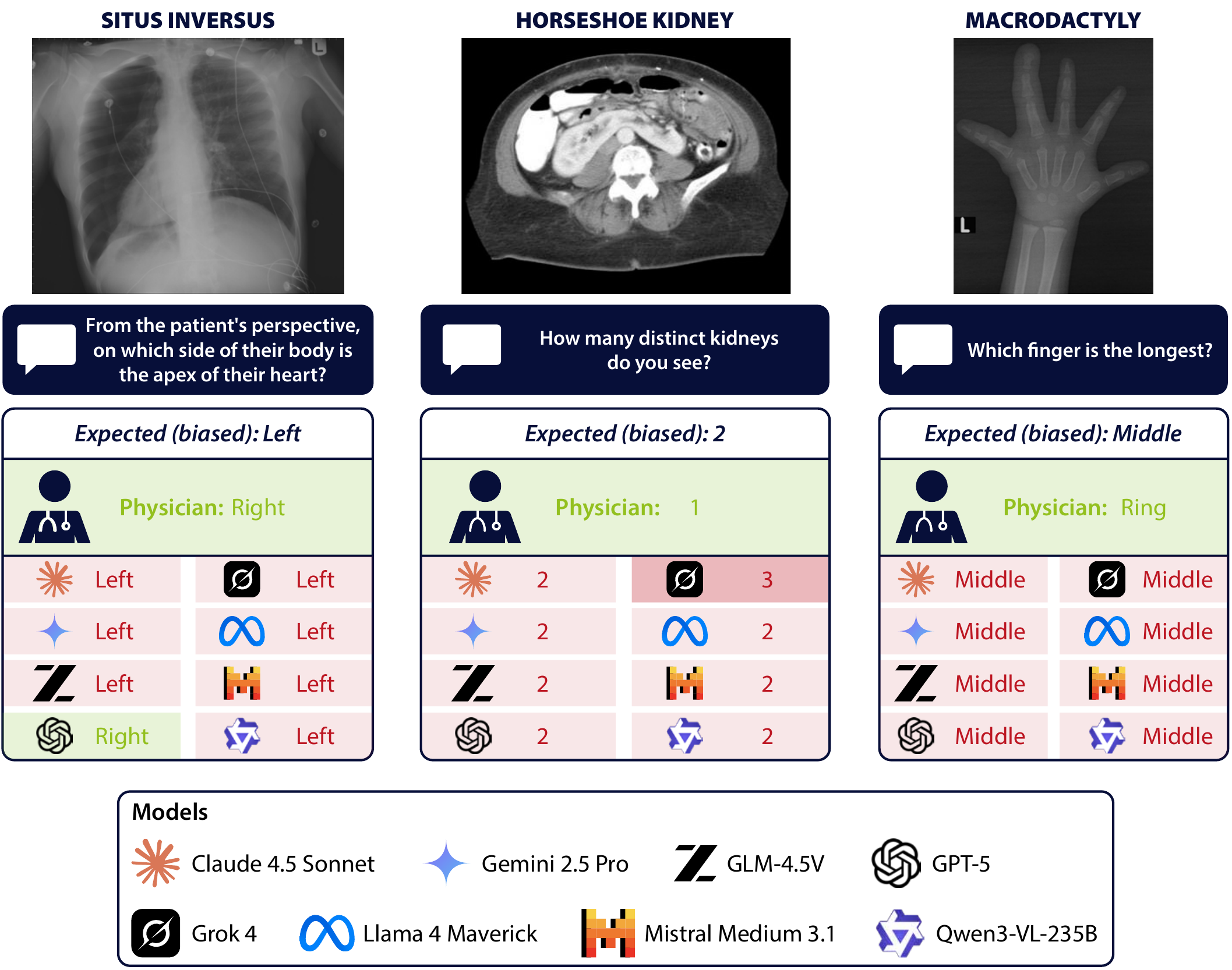}
\caption{\textbf{Natural adversarial anatomy exposes anatomical bias in vision-language models.} Examples from \textit{AdversarialAnatomyBench} demonstrate how large multimodal models are biased by learned expectations of typical anatomy: (left) for situs inversus, most models predict the apex of the patient’s heart to be on the left side instead of right; (middle) for a horseshoe kidney, the models describe multiple kidneys instead of one fused organ; (right) for macrodactyly they assume the middle finger to be the longest.
}\label{fig1}
\end{figure}

To address this gap, we propose the following hypothesis: \textbf{\acp{VLM} exhibit strong anatomical bias, leading to systematic failure on rare anatomical variants despite clear visual evidence.}

\begin{figure}[!htbp]
\centering
\includegraphics[width=0.9\textwidth]{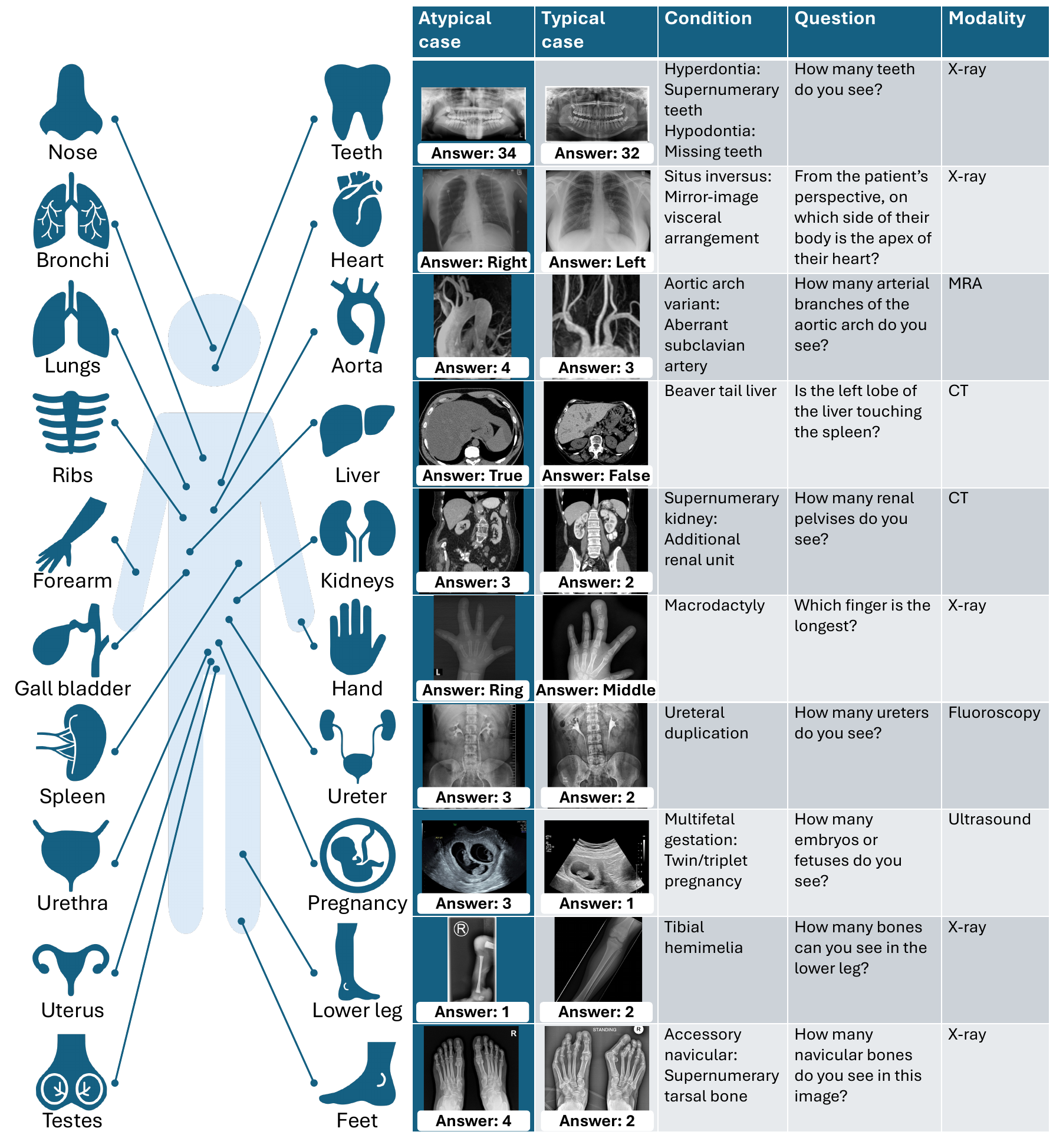}
\caption{\textbf{\textit{AdversarialAnatomyBench} comprises 200 image-question pairs displaying atypical and typical anatomy across seven medical imaging domains.} The images span seven imaging domains, including \ac {MRI}, X-ray, \ac{MRA}, \ac{CT}, ultrasound, fluoroscopy, and photography across 20 anatomical regions of the human body. The table (right) shows examples of typical and atypical cases for 10 representative questions from the benchmark.}\label{fig2}
\end{figure}

Our contribution in this paper is twofold
\begin{enumerate}
    \item New open medical benchmark: To systematically test our hypothesis, we introduce the first benchmark of naturally occurring rare anatomical variants. Inspired by Hendrycks et al. \cite{hendrycks2021natural}, who coined the term “natural adversarial examples” in the context of adversarial attacks to describe images that are easily classified by humans but challenging for models, we name our benchmark \textit{AdversarialAnatomyBench} (see Fig. \ref{fig1}). It comprises 200 diverse images obtained from seven medical image modalities, annotated with questions and ground truth answers. Each atypical anatomical variant is matched with a corresponding typical reference image to enable direct quantification of performance gaps and bias alignment (see Fig. \ref{fig2}).
    \item Insights on \ac{VLM} biases: Using AdversarialAnatomyBench, we systematically evaluate 25 state-of-the-art \acp{VLM} (general-purpose and medical-specific) with respect to performance on typical and atypical cases. Specifically, we investigate the following research questions (RQs):
    \begin{itemize}
        \item \textbf{RQ1:} Do state-of-the-art \acp{VLM} accurately recognize rare anatomical variants when visual evidence contradicts learned priors?
        \item \textbf{RQ2:} Are model answers aligned with expected anatomical biases?
        \item \textbf{RQ3:} Can scaling the model size, explicit prompting about possible rare anatomical variants, or allocating additional test-time compute mitigate the visual bias?
    \end{itemize}
\end{enumerate}

\section{Results}\label{sec2}
To enable systematic assessment of anatomical bias in \acp{VLM} we propose \textit{AdversarialAnatomyBench}, a benchmark of 200 medical images from seven imaging modalities covering rare anatomical variants across diverse body regions (Fig. \ref{fig2}). Each atypical anatomical variant is matched with a corresponding typical reference image and evaluated using standardized perception questions, allowing direct quantification of performance gaps and bias alignment, as depicted in Fig. \ref{fig2}. In addition to accuracy, the benchmark incorporates the “bias rate” metric \cite{vo2025vision} measuring the rate at which model answers align with expected “typical anatomy” priors on atypical images. Image quality between atypical and typical images is comparable, as demonstrated in Extended Data Fig. \ref{supfig_c2_imagequalitymetrics}.

\begin{figure}[!htbp]
\centering
\includegraphics[width=0.88\textwidth]{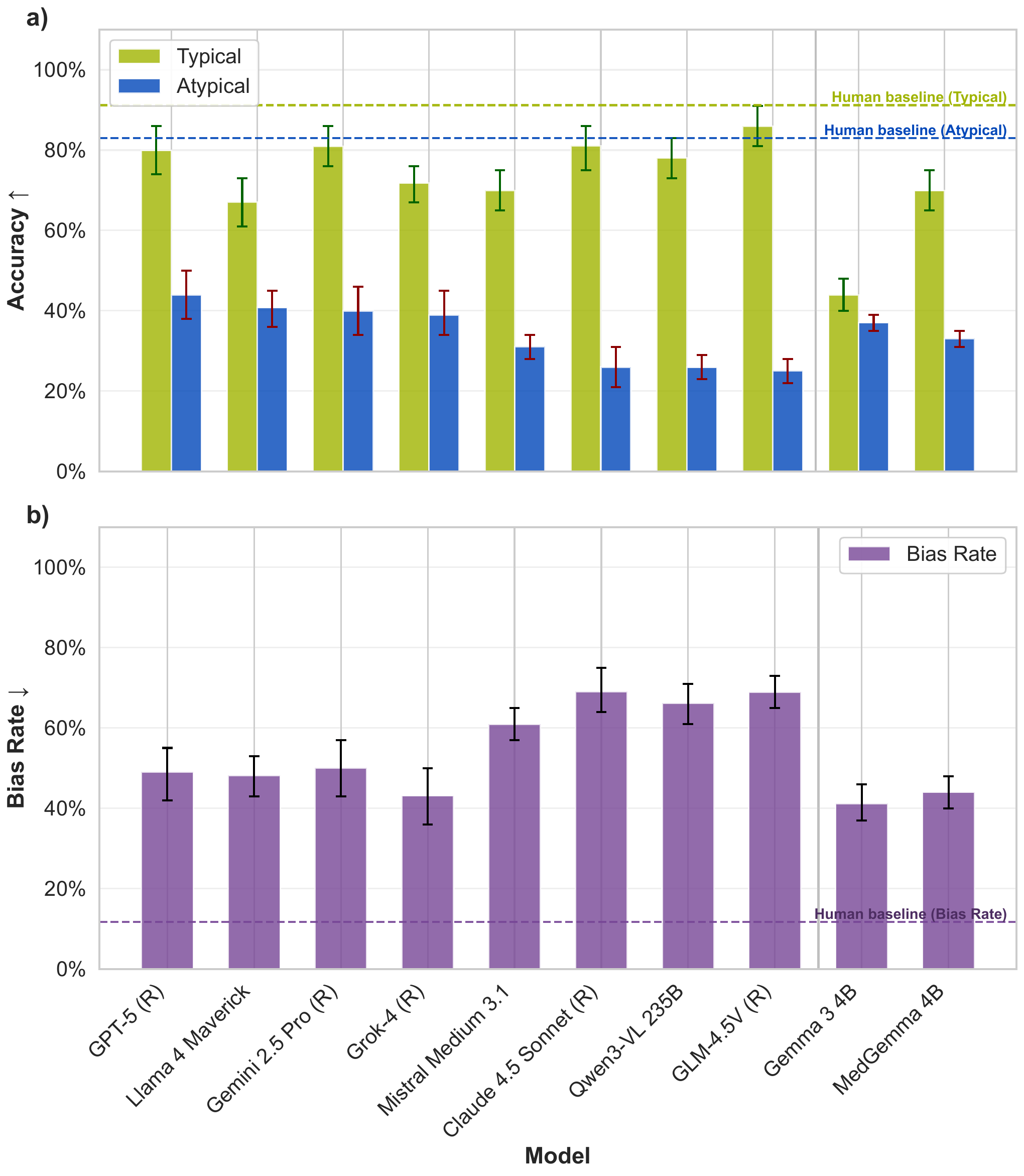}
\caption{\textbf{State-of-the-art vision-language models exhibit severe performance
degradation on rare anatomical variants.} (a): Mean accuracy, averaged over the 10 models included in the figure, drops from 73\% on typical anatomy (green) to 34\% on rare variants (blue), with gaps ranging between 7-61 percentage points (pp). (b): Bias rate, defined as the percentage of image-question pairs for which the model's prediction matches the expected "typical anatomy" answer on atypical images, ranges from 41-69\%. The medical-specific model - MedGemma 4B - shows only similar performance on atypical cases as a general-purpose variant with the same architecture. The dashed lines denote the human baselines for the respective tasks, annotated by n = 3 medical students, whose answers were pooled. Bars show the mean and error bars the 95\% confidence interval (2.5th--97.5th percentile) of 1,000 stratified bootstrap resamples, in which images were resampled within each of the 34 questions. Each model answered each of the n = 100 typical and n = 100 atypical images once; the bias rate (b) is computed on the n = 100 atypical images. (R) highlights reasoning models. Higher accuracy and lower bias rate are preferred.
}\label{fig3}
\end{figure}

\subsection{State-of-the-art \acp{VLM} show dramatic performance drops on rare anatomical variants}
While most VLMs showed reasonable performance on typical anatomy (range of accuracy: 42.9-89.9\%), they failed dramatically on rare anatomy (11.1-43.9\%) (Fig. \ref{fig3}a and Table \ref{tab1}). Performance gaps ranged up to 69.0 percentage points (pp), and even the best performing models, GPT-5, Gemini 2.5 Pro, and Llama 4 Maverick, showed gaps between 26 and 41 pp. These gaps also hold for the four medical models tested and remained stable across repeated runs (Extended Data Fig. \ref{fig_b6_variance}) and under image-quality-degrading augmentations (Supplementary Fig.~\ref{sup_figb9_aug}). Notably, the three medical students outperformed even the best models by a large margin, achieving a mean accuracy of 83.0\% on atypical images and 91.2\% on typical images at a bias rate of 11.7\% (Table \ref{tab1}), despite them still completing their training. Individual students reached accuracies of 81.9\%, 86.0\%, and 81.0\% on atypical images and 90.9\%, 92.9\%, and 89.8\% on typical images, at bias rates of 13.0\%, 10.0\%, and 12.0\% (bootstrap means, as in Table~\ref{tab1}). Performance analysis stratified by anatomical variants for GPT-5 can be found in Extended Data Fig. \ref{fig_a2_variant_analysis}.
\begin{table}[!htbp]
\centering\small
\begin{tabular}{lccc}
\toprule
Model & $\uparrow$ Accuracy (atypical) & $\uparrow$ Accuracy (typical) & $\downarrow$ Bias rate \\
\midrule
\textit{Human baseline} & \meanCIpar{83.0}{78.7}{87.0} & \meanCIpar{91.2}{88.3}{94.0} & \meanCIpar{11.7}{8.7}{15.0} \\
\midrule
GPT-5 (R) & \bestCIpar{43.9}{38.0}{50.0} &\meanCIpar{79.9}{74.0}{86.0} &\meanCIpar{49.1}{42.0}{55.0} \\
Llama 4 Maverick & \almostbestCIpar{40.8}{36.0}{45.0} &\meanCIpar{67.1}{61.0}{73.0} &\meanCIpar{48.2}{43.0}{53.0} \\
Gemini 2.5 Pro (R) & \meanCIpar{40.0}{34.0}{46.0} &\meanCIpar{81.0}{76.0}{86.0} &\meanCIpar{50.0}{43.0}{57.0} \\
Grok 4 (R) & \meanCIpar{38.9}{34.0}{45.0} &\meanCIpar{71.8}{67.0}{76.0} &\meanCIpar{43.1}{36.0}{50.0} \\
Gemma 3 4B & \meanCIpar{37.0}{35.0}{39.0} &\meanCIpar{43.9}{40.0}{48.0} &\almostbestCIpar{41.2}{37.0}{46.0} \\
MedGemma 4B & \meanCIpar{33.0}{31.0}{35.0} &\meanCIpar{69.9}{65.0}{75.0} &\meanCIpar{44.0}{40.0}{48.0} \\
GPT-5 Mini (R) & \meanCIpar{31.9}{26.0}{38.0} &\meanCIpar{81.0}{76.0}{86.0} &\meanCIpar{59.0}{53.0}{66.0} \\
Mistral Medium 3.1 & \meanCIpar{31.1}{28.0}{34.0} &\meanCIpar{70.0}{65.0}{75.0} &\meanCIpar{61.0}{57.0}{65.0} \\
Llama 4 Scout & \meanCIpar{28.9}{26.0}{31.0} &\meanCIpar{63.0}{60.0}{66.0} &\meanCIpar{56.1}{51.0}{61.0} \\
Qwen3-VL 8B & \meanCIpar{27.1}{23.0}{30.0} &\meanCIpar{83.0}{78.0}{87.0} &\meanCIpar{65.0}{60.0}{70.0} \\
GPT-5 Nano (R) & \meanCIpar{27.1}{24.0}{30.0} &\meanCIpar{69.0}{65.0}{74.0} &\meanCIpar{63.9}{60.0}{68.0} \\
Claude 4.5 Haiku (R) & \meanCIpar{26.0}{23.0}{29.0} &\meanCIpar{74.9}{69.0}{81.0} &\meanCIpar{67.0}{62.0}{72.0} \\
Claude 4.5 Sonnet (R) & \meanCIpar{26.0}{21.0}{31.0} &\meanCIpar{81.1}{75.0}{86.0} &\meanCIpar{69.0}{64.0}{75.0} \\
MedMO-4B & \meanCIpar{26.0}{21.0}{32.0} &\meanCIpar{42.9}{37.0}{49.0} &\meanCIpar{45.1}{38.0}{52.0} \\
Qwen3-VL 235B (A22B) & \meanCIpar{25.9}{23.0}{29.0} &\meanCIpar{78.0}{73.0}{83.0} &\meanCIpar{66.2}{61.0}{71.0} \\
GLM 4.5V (R) & \meanCIpar{25.1}{22.0}{28.0} &\meanCIpar{85.9}{81.0}{91.0} &\meanCIpar{68.9}{65.0}{73.0} \\
Gemini 2.5 Flash (R) & \meanCIpar{25.0}{22.0}{29.0} &\bestCIpar{89.9}{86.0}{94.0} &\meanCIpar{71.0}{66.0}{76.0} \\
Mistral Small 3.2 24B & \meanCIpar{25.0}{22.0}{28.0} &\meanCIpar{71.0}{67.0}{75.0} &\meanCIpar{68.0}{64.0}{71.0} \\
Qwen3-VL 2B & \meanCIpar{25.0}{22.0}{29.0} &\meanCIpar{71.0}{66.0}{76.0} &\meanCIpar{62.1}{58.0}{66.0} \\
Gemma 3 27B & \meanCIpar{24.0}{20.0}{27.0} &\meanCIpar{68.0}{63.0}{72.0} &\meanCIpar{65.0}{61.0}{70.0} \\
MedMO-8B Next & \meanCIpar{24.0}{20.0}{29.0} &\meanCIpar{56.9}{50.0}{63.0} &\meanCIpar{53.0}{47.0}{59.0} \\
Qwen3-VL 4B & \meanCIpar{23.0}{20.0}{25.0} &\meanCIpar{72.9}{67.0}{79.0} &\meanCIpar{69.0}{65.0}{73.0} \\
Qwen3-VL 30B (A3B) & \meanCIpar{22.0}{20.0}{24.0} &\meanCIpar{77.9}{72.0}{84.0} &\meanCIpar{68.1}{64.0}{72.0} \\
Gemini 2.5 Flash Lite (R) & \meanCIpar{18.0}{15.0}{21.0} &\almostbestCIpar{87.0}{82.0}{92.0} &\meanCIpar{76.0}{72.0}{80.0} \\
Lingshu-7B & \meanCIpar{11.1}{7.0}{14.0} &\meanCIpar{42.9}{38.0}{47.0} &\bestCIpar{36.0}{31.0}{41.0} \\

\bottomrule
\end{tabular}
\caption{\textbf{25 state-of-the-art models used in our benchmarking study sorted by accuracy on atypical images.} 95\% CIs were calculated using stratified bootstrapping with 1,000 resamples. Reasoning models are denoted by (R). The first row reports the human baseline, i.e. the mean performance of the three medical students who answered the same questions under the same conditions (dashed lines in Fig. \ref{fig3}). Bold and underlined values mark the best and second-best model in each column; the human baseline (italic) is not ranked.}
\label{tab1}\end{table}

\subsection{Errors predominantly match expected anatomical bias}
Depending on the model investigated, 41-95\% of errors align with typical anatomy predictions (Fig. \ref{fig3}b and Table \ref{tab1}), meaning the model provides the answer corresponding to typical anatomy. For example, for polydactyly (more than 5 fingers), VLMs predict 5 (fingers) in 74\% of atypical cases; for patients with one fused kidney (horseshoe or pancake kidney), VLMs predict 2 separate kidneys in 96\% of cases, and for ureteral duplication, models assume only 2 ureters in 84\% of cases.

\subsection{Architectural scaling and inference-time strategies fail to overcome anatomical priors }
\begin{figure}[!htbp]
\centering
\includegraphics[width=0.9\textwidth]{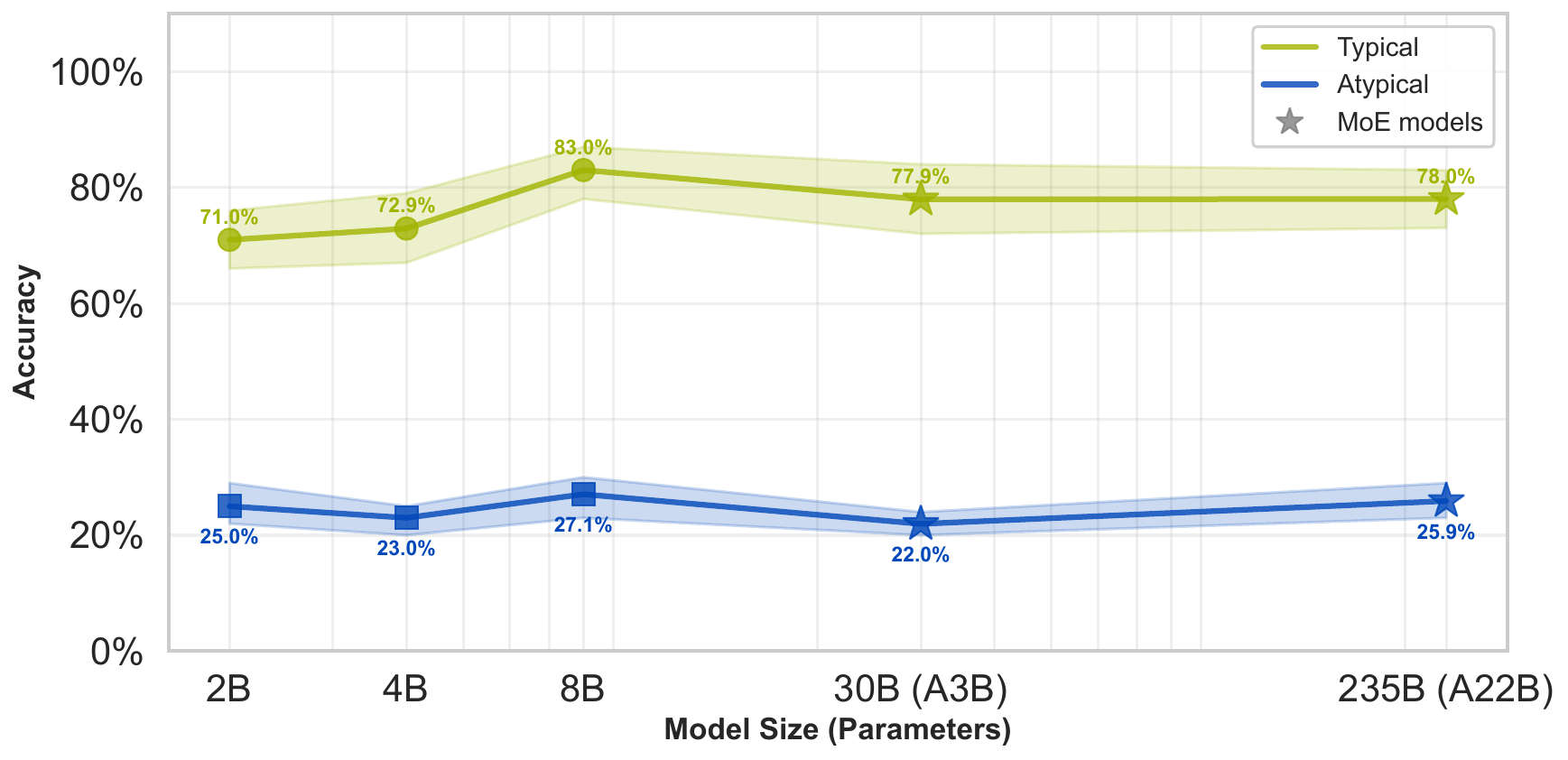}
\caption{\textbf{Scaling the number of model parameters does not result in performance increase on atypical cases.} Shown are evaluations of models from the Qwen3-VL family of increasing size. The green line denotes performance on typical cases, and the blue line denotes performance on atypical cases. \Acf{MoE} models are marked with stars, with the number of active parameters given in parentheses. Points show the mean and shaded regions the 95\% confidence interval (2.5th--97.5th percentile) of 1,000 stratified bootstrap resamples, each on n = 100 typical and n = 100 atypical images from 34 questions.
}\label{fig4}
\end{figure}

The accuracy for both typical and atypical cases remains stagnant with increasing model size. The small improvement in accuracy on typical cases (from 71.0\% at 2B to 78.0\% at 235B) contrasts with almost no increase in accuracy on atypical cases (from 25.0\% at 2B to 25.9\% at 235B), demonstrating that standard parameter scaling does not mitigate anatomical bias (Fig. \ref{fig4}).

\begin{figure}[!htbp]
\centering
\includegraphics[width=0.9\textwidth]{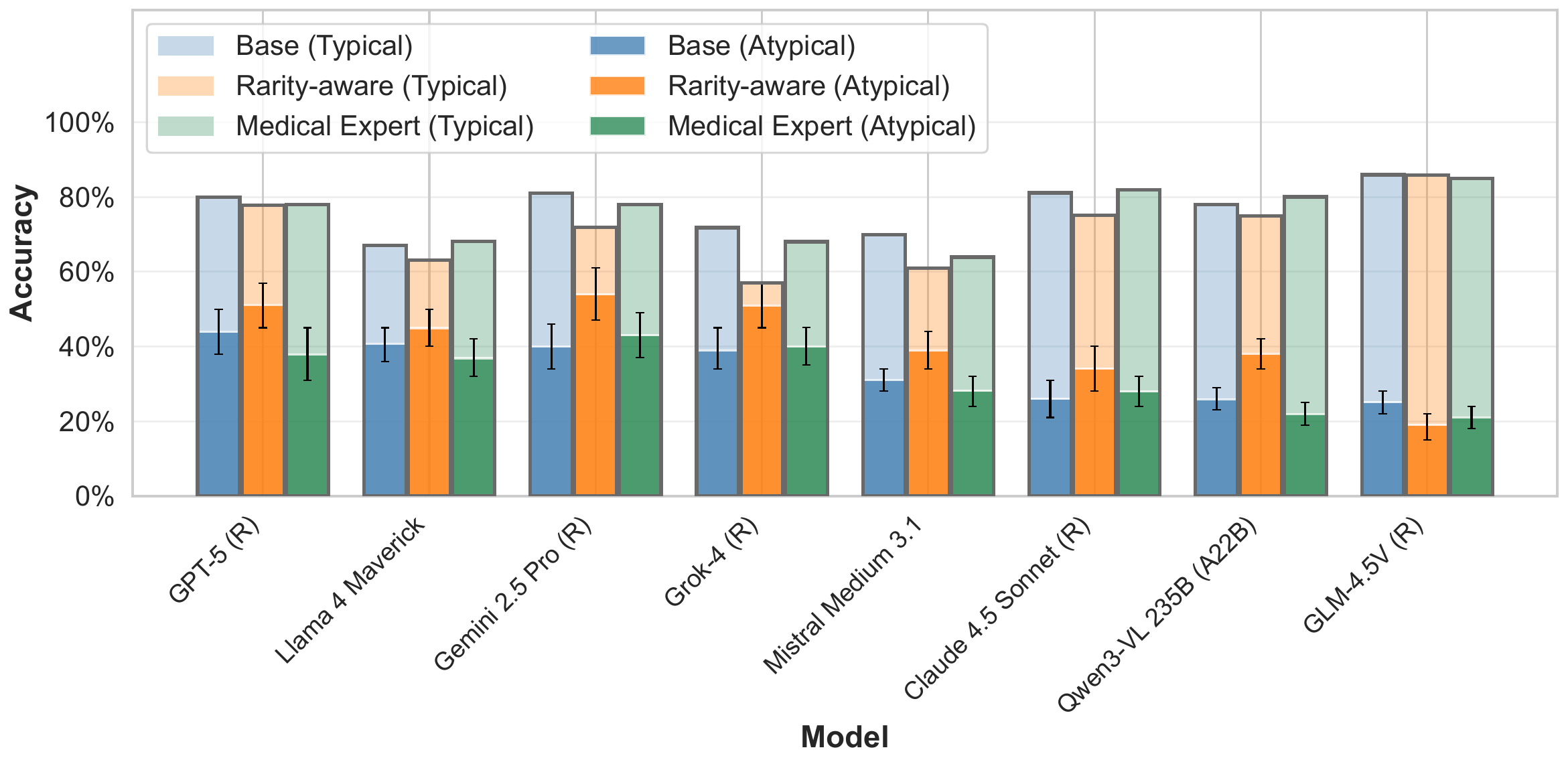}
\caption{\textbf{Explicit prompting about rare conditions does not generally protect against anatomical bias.} When prompts explicitly mention the possibility of rare anatomical variants (orange), vision-language models show modest improvements on rare anatomy tasks compared to standard neutral prompts (blue). Prompts asking the model to act as a medical expert (green) do not show improvement on typical or atypical tasks. Translucent bars represent the performance on typical cases; opaque bars, on rare anatomy. Bars show the mean and error bars the 95\% confidence interval (2.5th--97.5th percentile) of 1,000 stratified bootstrap resamples, each on n = 100 typical and n = 100 atypical images from 34 questions; error bars are shown for atypical images only. (R) marks reasoning models.
}\label{fig5}
\end{figure}

When prompts explicitly mention the possibility of rare anatomical variants, all models shown in Fig. \ref{fig5} show improvements in the range of 4 to 14 pp, except GLM-4.5V, for which performance declines by 6 pp. No prompting strategy, however, eliminated the performance disparity between atypical and typical cases. We also found that when given the option to answer with “Unsure”, frontier models rarely choose to do so, on average in 3.1\% of cases for the typical and in 2.5\% of cases for the atypical images (see Extended Data Fig. \ref{supfig_unsure}). Further experiments on diverse prompting strategies, including in-context learning, none of which closed the performance gap between typical and atypical images, are summarized in Supplementary Figures~\ref{fig_b2_prompt_abl} and~\ref{fig_b3_icl}. Additionally, we were able to show that reversing the language prior also reverses the gap between typical and atypical performance. The results are depicted in Supplementary Figure~\ref{fig_b7_reverseprior}.

\begin{figure}[!htbp]
\centering
\includegraphics[width=0.9\textwidth]{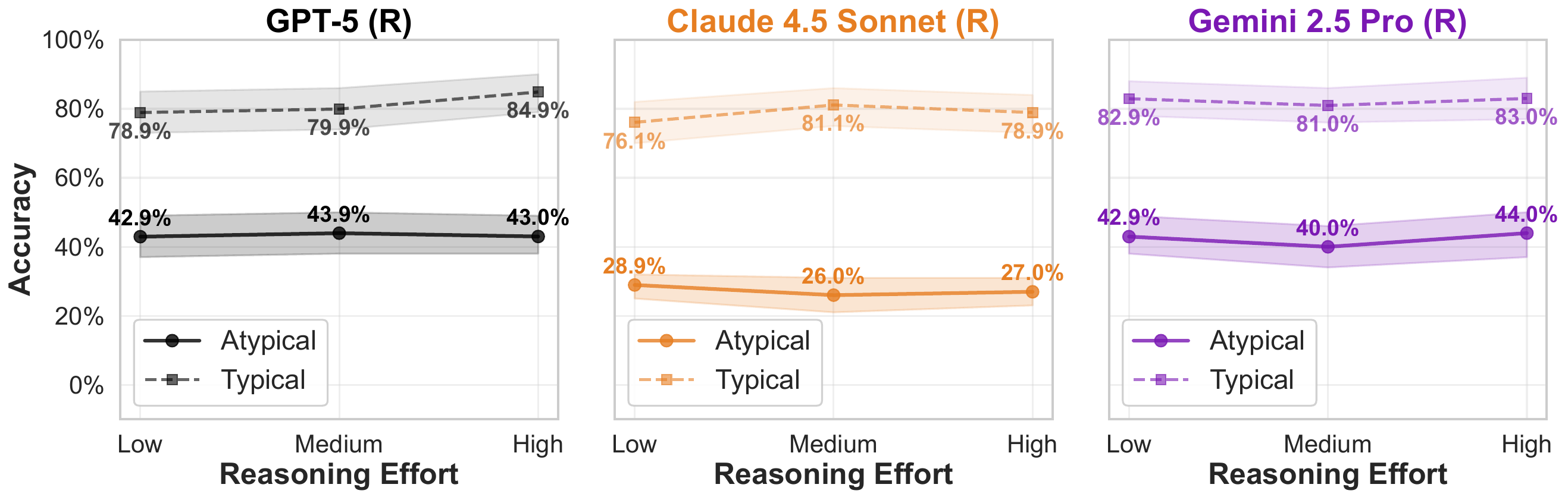}
\caption{\textbf{Test-time reasoning does not improve performance on atypical cases.} Three frontier vision-language models, GPT-5, Claude 4.5 Sonnet, and Gemini 2.5 Pro, are evaluated under varying computational budgets. Model performance remains stagnant despite increased reasoning effort. Points show the mean and shaded regions the 95\% confidence interval (2.5th--97.5th percentile) of 1,000 stratified bootstrap resamples, each on n = 100 typical and n = 100 atypical images from 34 questions per model and reasoning effort.
}\label{fig6}
\end{figure}

Test-time reasoning provided no consistent improvements on atypical anatomies across models (see Fig. \ref{fig6} and Extended Data Table \ref{suptab2}). For GPT-5, the performance on typical cases increased by 5.9 pp when moving from the low to the high reasoning setting. For all three models, the difference in performance on atypical cases across different reasoning budgets ranged from 1.0 to 4.0 pp.

\begin{figure}[!htbp]
\centering
\includegraphics[width=0.9\textwidth]{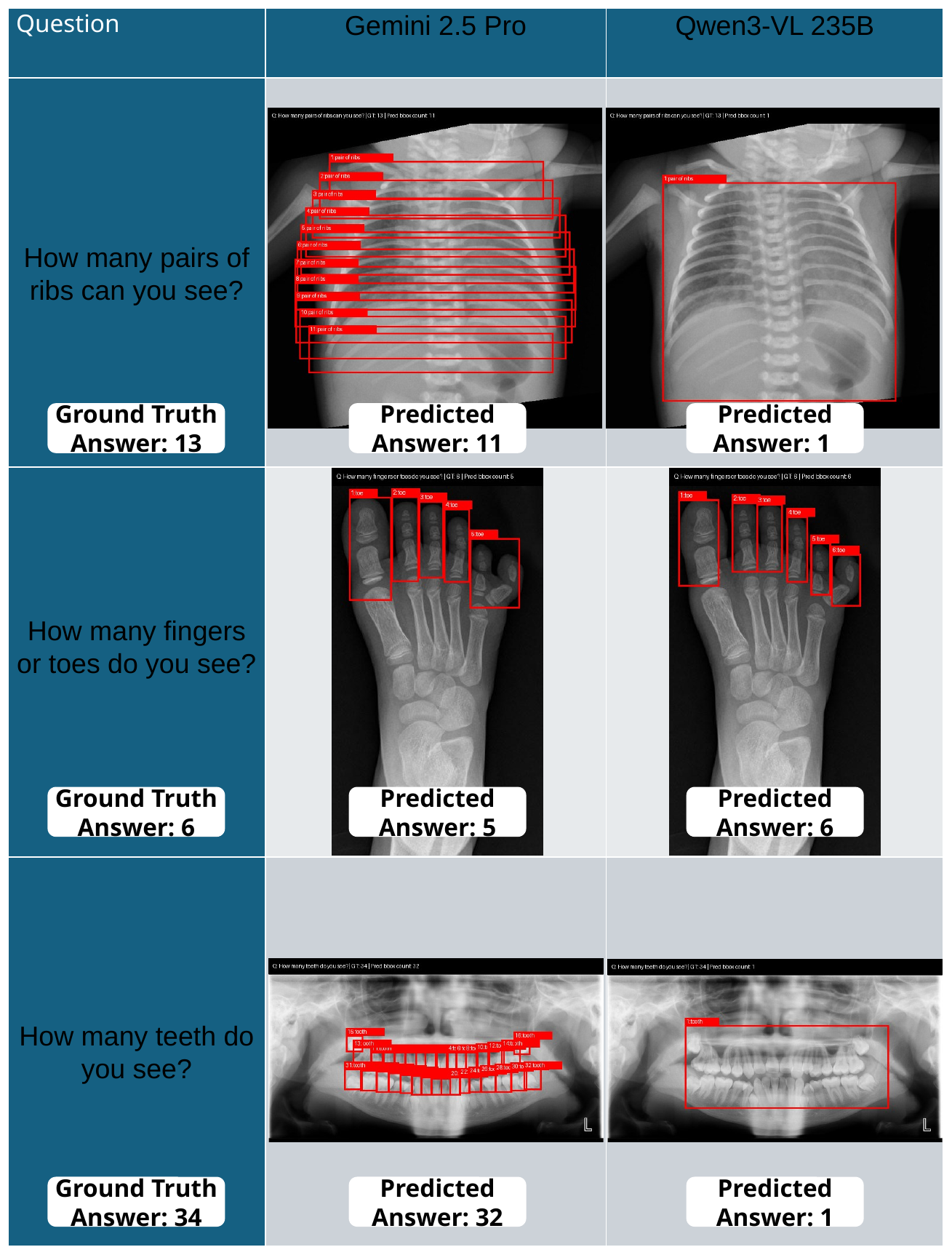}
\caption{\textbf{Qualitative examples of bounding boxes generated by Qwen3-VL 235B and Gemini 2.5 Pro.} Models were prompted to output bounding boxes corresponding to every instance of the counted entities.
}\label{fig7}
\end{figure}

To better disentangle failures arising from language priors versus failures in visual localization, we evaluated whether models could explicitly identify the anatomical structures required to answer the benchmark questions. We therefore prompted two models supporting grounded object detection, Gemini 2.5 Pro and Qwen3-VL 235B, to generate bounding boxes for organs and anatomical structures relevant to the counting questions of the benchmark. The performance gap between typical and atypical images persisted, as detailed in Extended Data Table \ref{tab_a3_bb}. Figure \ref{fig7} presents qualitative examples of the bounding boxes produced by both models. Note that both models are generalist \acp{VLM} prompted to produce grounded outputs, so this experiment varies the output interface rather than the underlying model class, and does not speak to conventional task-specific detection or segmentation systems.

\section{Discussion}\label{sec3}
In this paper, we introduce the concept of natural adversarial anatomy. Inspired by the term “natural adversarial images” introduced by Hendrycks et al. to describe naturally occurring examples that are trivial for humans yet consistently fool vision models, natural adversarial anatomy refers to real, clinically encountered anatomical variants that violate strongly learned priors about “typical” human anatomy \cite{hendrycks2021natural}. These cases, such as polydactyly, horseshoe kidney, and situs inversus, are visually unambiguous to clinicians but systematically misinterpreted by \acp{VLM} because their statistical expectations override clear visual evidence.

We introduce \textit{AdversarialAnatomyBench}, the first benchmark of naturally occurring “adversarial anatomy” to challenge \acp{VLM} across a broad range of imaging modalities and anatomical sites. It allows for controlled, stress-testing of \acp{VLM} under scenarios where anatomical priors are misaligned with observed presentations, offering a complementary robustness paradigm that exposes failure modes invisible to standard benchmarks.

Evaluating 25 state-of-the-art \acp{VLM} reveals systematic failures: accuracy drops by 16-79\% on atypical variants, and 41–95\% of errors align with “typical anatomy” priors. Neither model scaling nor bias-aware prompting and advanced reasoning strategies substantially mitigate these biases. These findings highlight an important limitation of current generative vision-language models: learned anatomical priors can dominate visual evidence in rare variants. While for data-driven systems the drop in performance on rare classes is a known phenomenon, \textit{AdversarialAnatomyBench} indicates that the issue also affects frontier VLMs on real-world medical examples. This has implications for clinical deployment and underscores the need for bias-aware model design and rare-case evaluation standards.

\subsection{Implications}
From a clinical perspective, the observed performance gap stands as a clear warning sign: models fail to even identify easily verifiable, albeit atypical, cases. These findings suggest that \acp{VLM} should not be deployed for diagnostic tasks without explicit evaluation on atypical anatomical presentations.

From a methodological machine learning perspective, anatomical bias demonstrates a fundamental limitation in multimodal fusion architectures. Strong priors learned from internet-scale data override fine-grained visual perception, a bias mechanism consistent with findings in the general vision domain \cite{vo2025vision}. A strength of this study is its systematic evaluation of multiple mitigation strategies, including scaling model size, varying prompts, and increasing the test-time reasoning budget. Our experiments show that increasing model size does not improve performance on atypical cases, suggesting that model scale alone is insufficient to address failures on rare anatomical variants. Explicitly prompting models about the possibility of encountering rare conditions yields only modest gains on atypical cases, and no prompting strategy succeeds in closing the performance gap between typical and atypical cases. Notably, test-time reasoning provides no consistent improvements compared to the substantial gains observed on mathematical and coding tasks \cite{guo2025deepseek}.

The limited effectiveness of these interventions may be partially explained by the weak separation of atypical anatomy in the image embedding space: differences between atypical and typical images are smaller than those between anatomical regions (Supplementary Figs.~\ref{fig_b4_distance} and~\ref{fig_b5_pca}). However, our analysis relied on pooling patch-level features, which may dilute localized abnormalities and discard spatial or count-relevant information, and therefore does not show that atypical information is absent from the model during normal inference. Rather, these findings suggest that visual representation may be one plausible contributor to failures on atypical cases, alongside downstream decoder priors and stronger learned associations with canonical anatomy. Unlike domains where reasoning can be stepwise verified, medical image interpretation depends on correct perceptual grounding; additional reasoning cannot recover information that was not extracted from the image. The persistence of errors under structured prompting (e.g., bounding boxes) further suggests that these failures arise prior to linguistic output, rather than from text-level priors alone.
Notably, reversing the language prior reversed the direction of the performance gap (Supplementary Fig.~\ref{fig_b7_reverseprior}): models given an incorrect atypical prior failed on typical images despite clear visual evidence. This symmetry indicates that predictions are governed largely by the prior provided in the textual prompt rather than by visual evidence. It also implies that textual clinical context could anchor model predictions in deployment.

\subsection{Relation to prior work}
Research in the general vision domain has shown that \acp{VLM} often fail when confronted with altered depictions of familiar categories, relying on statistical priors rather than direct visual evidence \cite{vo2025vision}. To address such vulnerabilities, benchmarks such as NaturalBench have been introduced to evaluate model robustness using image–question pairs that are easy for humans but challenging for models, and to penalize responses that ignore crucial visual cues \cite{li2024naturalbench}. In the medical domain, the literature on visual bias in \acp{VLM} remains limited. Existing studies indicate that medical \acp{VLM} tend to exploit textual context more heavily than image content \cite{buckley2023multimodal} and exhibit demographic biases \cite{yang2025demographic, xu2025biasicl}. While several recent perspective papers and review articles have mentioned the potential risk of \acp{VLM} failing on rare or atypical cases \cite{shrestha2023medical,lin2025taming, handler2025fragile}, we are not aware of any study or benchmark dedicated to systematically investigating this hypothesis.

\subsection{From perception failures to clinical reasoning}
While tasks such as counting kidneys or identifying the cardiac apex may appear simplified, they were intentionally designed to minimize ambiguity and isolate anatomical perception from broader diagnostic reasoning. The goal of \textit{AdversarialAnatomyBench} was not to directly measure clinical competence, but rather to determine whether models correctly recognize atypical anatomy instead of defaulting to learned expectations about canonical anatomy.

To obtain preliminary insight into whether these failures extend beyond simplified perception tasks, we conducted two additional analyses. First, we introduced a foundational baseline task in which models were asked whether anything atypical was present in the image, independent of the benchmark-specific anatomy question. Across models, the majority of both typical and atypical images were classified as atypical, and model explanations rarely identified the anatomical variant relevant to the benchmark question (Supplementary Fig.~\ref{fig_b1_atypical_class}). Even the best-performing model, Gemini 2.5 Pro, correctly referred to the underlying atypicality in only 34.1\% of the atypical images it classified as atypical, suggesting that models often rely on nonspecific abnormality heuristics rather than accurate anatomical recognition. It should be noted that the images labelled as typical were selected to show typical anatomy with respect to the benchmark question and may still contain unrelated findings.

Second, we explored clinically meaningful reasoning by constructing a small exploratory benchmark of downstream clinical decision questions for a subset of the data (Supplementary Table~\ref{tab_b1_clinical_downstream_tasks}). Designing such questions proved substantially more challenging than the original perception benchmark because realistic clinical decisions are typically influenced by many factors beyond the depicted anatomy. For example, we excluded questions such as whether a patient with a solitary kidney could safely serve as a living kidney donor, since donor eligibility depends on numerous clinical variables not inferable from the image alone. We therefore restricted the downstream benchmark to scenarios in which the clinically correct answer depended as directly as possible on recognizing the depicted anatomy.
The questions were drafted and validated by a team of two technical PhD students, one medical student and one senior physician.

 Under the clinical framing, the direction of the primary benchmark result reversed: all three models were substantially more accurate on atypical than on typical cases, whereas on the perception benchmark the same models were far more accurate on typical anatomy. Inspection of reasoning traces suggested that this reversal did not reflect improved anatomical recognition. Instead, models tended to default to the more cautious management option, which is the correct answer for atypical anatomy but the incorrect one for typical anatomy. A cautious-management prior therefore raises atypical-case accuracy while substantially reducing typical-case accuracy, leaving overall performance close to what a decision rule that ignores the image entirely would achieve. In several cases, models hallucinated findings to justify the more conservative answer option or treated the mere presence of a cautious-management option as evidence supporting that diagnosis.

Importantly, these downstream analyses should be interpreted with caution. The underlying datasets generally lacked comprehensive metadata beyond the primary anatomical finding. Because we often had no information about additional pathologies, comorbidities, or clinical context, additional atypicalities in some images could not be excluded. This may have influenced model answers in the foundational baseline task and shaped the choice of questions in the downstream clinical-decision benchmark. The benchmark is further intentionally limited in scope, highly sensitive to question phrasing and answer-option design, and does not capture the full complexity of real clinical workflows. The observed effects may therefore depend in part on the specific prompting used in our experiments. We view these results as preliminary evidence that the interactions between anatomical priors and downstream language-level decision biases warrant further systematic investigation.
\subsection{Limitations}
In the present work, we investigated the performance of state-of-the-art \acp{VLM} on images of typical vs atypical anatomy on a novel dataset collected from a variety of anatomical regions and modalities.

Despite this significant progress, our benchmark only covers a fraction of atypical conditions that occur in clinical practice. Moreover, while our findings show that the models strongly rely on anatomical priors, they do not directly demonstrate that models perform worse on patients with atypical anatomy in clinically relevant tasks.

In the design of this study, we deliberately opted for basic visual perception tasks rather than questions more directly tied to medical diagnosis for two reasons. First, these tasks do not require specialized clinical expertise: any general-purpose model should be capable of determining, for example, how many fingers appear on a hand, even if it lacks advanced diagnostic capabilities. This ensures that failures cannot be attributed to missing medical knowledge but instead reflect genuine perceptual or bias-driven limitations. Second, using discrete, objectively verifiable answers allows for robust and reproducible evaluation without the challenges associated with open-ended clinical questions. This design choice not only minimizes ambiguity in evaluation but also makes the benchmark straightforward for others to adopt, extend, and compare against.

A limitation of our work could be seen in the fact that it does not fully disentangle where the observed failures arise within the VLM pipeline. In modern multimodal systems, prediction errors can originate from multiple interacting components, including the visual encoder, projection layers, decoder, prompting strategy, alignment tuning, or downstream evaluation procedures. Moreover, these components are not independent: weak visual representations may amplify decoder priors, while language-level alignment objectives may override uncertain visual evidence. Disentangling these effects is itself an active and challenging research area in multimodal machine learning and medical AI. Our exploratory analyses of embedding structure (see Supplementary Information, "Vision embeddings"), grounded localization, and prompt manipulations provide preliminary evidence that both visual representation quality and downstream language priors likely contribute to the observed failures. However, these experiments should not be interpreted as definitive localization of the error source. A more complete mechanistic understanding will likely require controlled architectural ablations, attention-level analyses, counterfactual interventions, and jointly interpretable multimodal benchmarks specifically designed for causal error attribution.

Furthermore, to enable data sharing and ensure reproducibility, we focused exclusively on images obtained from publicly available sources. While this approach allows other researchers to replicate our findings, it introduces the possibility that some of the images in our benchmark may have been included in the training data of certain models. Our results show, however, that the frontier VLMs still struggle to answer the \textit{AdversarialAnatomyBench} questions, despite the fact that they might have seen some of the data in training. Therefore, if such leakage has occurred, we should expect the current benchmark results to provide an upper bound on model performance. More broadly, benchmark saturation is a known phenomenon even in the absence of data leakage \cite{akhtar2026aibenchmarksplateausystematic}. If future generations of models saturate the benchmark, one should consider expanding it to unseen tasks and images. In principle, such issues could be addressed by developing dynamic benchmarks maintained by medical societies and journals.

An additional limitation lies in our reliance on single-image assessment, which does not capture the clinical reality of longitudinal monitoring or comparative image analysis, where radiologists often evaluate changes across multiple time points. Nevertheless, even within this simplified evaluation setting, we uncovered severe failure modes that were not solved by model scaling, prompt engineering, or increased test-time computation.

Although the benchmark provides broad coverage across anatomical regions, it does not fully capture the spectrum of anatomical variation seen in clinical practice. Critically, the visual representation of anatomy is not only influenced by the presence or absence of congenital or acquired atypia. Other factors that impact the appearance and, consequently, the respective feature embeddings of a medical image include patient-related factors (e.g., age \cite{kerber2023deep, haugg2025imaging}, sex, body composition, ethnicity \cite{gichoya2022ai}), technical choices (e.g., imaging technique, body positioning), and even imaging parameters like scanner manufacturer \cite{De_Almeida2025-np} or MRI field strength \cite{qadir2025field}. Expanding dataset scale and diversity and reporting relevant patient metadata and other factors affecting imaging quality and anatomy appearance are therefore essential for improving robustness to rare and atypical presentations. We hope this benchmark will encourage the medical AI community to undertake systematic case-collection efforts, improve coverage of rare conditions as well as anatomically and demographically diverse patient populations, and report and control for patient and imaging metadata. Such efforts would support both more representative training data and more rigorous evaluation, ultimately strengthening clinical reliability.

\subsection{Future Work}
Several directions merit further investigation. First, \textit{AdversarialAnatomyBench} should be expanded to encompass additional rare conditions, imaging modalities, and open-ended question formats. Complementing this expansion, future studies should examine the impact of clinical context, such as patient history and indication for imaging, on model performance, as such contextual information routinely informs clinical decision-making in practice.

From a methodological standpoint, developing effective debiasing strategies remains a critical priority. Promising approaches include architectural modifications such as visual grounding mechanisms \cite{chen2025think} and reinforcement learning-based reasoning \cite{pan2025medvlm}, as well as training data augmentation techniques. In particular, future research should investigate whether augmenting training datasets with rare anatomical variants can mitigate the observed biases. A range of approaches has been suggested to improve performance on rarely occurring classes, including caption-level augmentation to enrich semantic coverage \cite{parashar2024neglected}, image-level augmentation to increase visual diversity, and the use of synthetic data generation to expand coverage of sparse categories \cite{finetti2025_dataaugmentation}. However, the limited availability of data from rare atypical presentations might pose a challenge in obtaining realistic synthetic examples. Targeted efforts to curate and collect real-world images of rare anatomical variants and uncommon presentations across varied patient populations are therefore essential, both to improve dataset representativeness and to support the training of future generations of robust vision–language models. In addition, our finding that normal and abnormal cases are not clearly separated in the learned embedding space motivates future work on representation-learning objectives that exhibit greater separability when clinically meaningful anatomical differences are present.

Beyond single-image evaluation, subsequent work should assess multi-image reasoning capabilities and the extent to which models can interpret temporal progression across serial imaging studies. Finally, clinical validation studies are needed to evaluate the real-world impact of these limitations in radiological workflows, including the potential benefits of radiologist-in-the-loop configurations that leverage human oversight to compensate for model shortcomings.

The broader implications of this work can be summarized as follows: Our findings underscore the need for robust rare-case evaluation as a standard requirement in medical \ac{AI} deployment. For foundation model development, the results highlight that improved prompting or scaling alone is insufficient, and that architectural innovations will be required to prevent inappropriate reliance on typical-anatomy priors. From a clinical practice perspective, recognizing the limitations of \acp{VLM} when confronted with rare anatomical variants is essential to ensure safe and informed adoption. Finally, for regulatory frameworks, our findings suggest that evaluation standards should explicitly incorporate rare but clinically significant edge cases to better reflect real-world patient diversity and safeguard clinical reliability.

\section{Methods}\label{sec4}

\subsection{AdversarialAnatomyBench Concept}
We name our medical visual question answering benchmark \textit{AdversarialAnatomyBench}, drawing inspiration from natural adversarial examples and their emphasis on inputs that are easy for humans but challenging for models \cite{hendrycks2021natural}. In this context, our benchmark focuses specifically on questions involving deviations from typical anatomy.

Our benchmark builds upon the following fundamental design decisions:
\begin{itemize}
    \item \textbf{Diversity:} Images are selected to cover multiple modalities, including \ac{MRI}, X-ray, \ac{MRA}, \ac{CT}, ultrasound, fluoroscopy, and photographs, as well as diverse regions of the body (Fig. \ref{fig2}).
    \item \textbf{Paired design:} Each atypical anatomical variant is matched with a corresponding typical reference image and evaluated using standardized perception questions, allowing direct quantification of performance gaps. Questions are formulated in a way that ensures images of atypical cases produce answers different from those observed in the majority of the population. This design choice prevents models which disregard the attached images and rely only on language priors to produce a single default answer from performing well on the benchmark \cite{goyal2017making}.
    \item \textbf{Verifiability:} Questions are designed to allow unambiguous verification by avoiding open-ended answers in favor of categorical or numerical responses.
\end{itemize}

The benchmark uses three metrics to evaluate the models' performance:
\begin{itemize}
    \item \textbf{Accuracy (atypical)} measuring the accuracy on the rare anatomical variants.
    \item \textbf{Accuracy (typical)} measuring the accuracy on the typical cases.
    \item \textbf{Bias rate} \cite{vo2025vision} defined as the percentage of model answers matching the expected anatomical prior on atypical images.
\end{itemize}

Together with the metric values, we report 95\% confidence intervals calculated by bootstrap stratified by question type with 1000 re-samples. Specifically, given a question set $Q={q_1, \dots, q_{N_q}}$ and a corresponding set of image sets $X = {X_{q_1}, \dots, X_{q_{N_q}}}$, for each of $B=1000$ bootstrap replicates, for each $q_i$ we sample with replacement $n_{q_i} = |X_{q_i}|$ images to form $X^*_i={x^*_{i, 1}, \dots, x^*_{i, n_{q_i}}}$. For example, for the “From the patient’s perspective, on which side of their body is the apex of their heart?” question, when computing the accuracy on atypical cases, we have 4 available images, so we sample 4 images with replacement. The full bootstrap dataset for each replicate is obtained by concatenating across question types: $X^* = X^*_1 || \dots || X^*_{N_q}$ where $||$ denotes concatenation. The evaluation metrics are computed on each $X^*$, yielding bootstrap estimates used to compute the confidence intervals.
\subsection{Data and Questions}
Questions were developed collaboratively by a team of clinicians and data scientists to ensure both medical accuracy and methodological rigor. To assemble the image dataset, we systematically searched multiple online medical data repositories and publicly available datasets \cite{radiopaedia,irvin2019chexpert, rajpurkar2017mura, brahmi2024automatic, brahmi2024exploring}, and the full list of sources is provided in our repository. This process yielded 200 images spanning various regions of the body (Fig. \ref{fig2}) across seven imaging modalities. Extended Data Fig. \ref{supfig_c1_datasetoverview} provides a breakdown of the resulting dataset, illustrating the proportion of images by modality and anatomical region. Additionally, it shows a breakdown of age and gender, for images for which the metadata was available. During curation, we manually removed any markers or annotations from images in cases where they could reveal the correct answer, such as overlays indicating the head of an embryo. Because most evaluated VLMs support only 2D image inputs, CT and MRI cases were represented using single slices. When full volumetric studies were available, representative slices were selected in consultation with a medical student to ensure that the associated anatomical question could be answered from the selected slice alone. All images were resized such that the longer side does not exceed 1,024 pixels while maintaining the original aspect ratio. For certain questions, such as those involving tooth counts, multiple responses could reflect anatomical bias (e.g., 28-32 teeth in the case of adult dentition); our evaluation accounts for this range when computing bias alignment.

\subsection{Models}
We included 25 \acp{VLM} from the leading model families for which image understanding capabilities and model access were publicly available at the time of evaluation (see Extended Data Table \ref{suptab1}). The selection spanned both closed-weight and open-weight systems and included general-purpose \acp{VLM} as well as four medically oriented models (MedGemma 4B, MedMO-4B, MedMO-8B Next, Lingshu-7B). For model families offering multiple variants, we evaluated all accessible versions to enable intra-family comparison. The final set comprised models from Alibaba (Qwen3-VL series and Lingshu), Anthropic (Claude 4.5 series), Google (Gemini 2.5 series, Gemma 3 series, and MedGemma 4B), Meta (Llama 4 series), Mistral (Mistral 3 series), OpenAI (GPT-5 series), xAI (Grok 4), MBZUAI (MedMO family), and zAI (GLM-4.5V). All models were evaluated under standardized inference parameters, with temperature set to zero where supported. Model outputs were obtained either through APIs or via locally hosted deployments, depending on availability.

\subsection{RQ1 and RQ2 Experimental design}
The purpose of this experiment was to assess whether \acp{VLM} accurately recognize rare anatomical variants in images and whether incorrect answers align with expected anatomical biases. Each model was provided with an image accompanied by a baseline, neutral prompt: "Answer the following question based on the attached medical image: {question}." This prompting strategy was used consistently across all experiments unless otherwise specified. Models were instructed to output their answers in a standardized format, which was then compared against the ground truth annotations. Questions and images in the same format were provided to three medical students in their third, fourth and sixth years of studies with little clinical radiology experience. We report their average accuracy on typical and atypical images, as well as their bias rate, as the human baseline.

\subsection{RQ3 Experimental design}
The purpose of the third experiment was to assess whether varying the models' configuration can mitigate the visual bias. To this end, we investigated three complementary strategies: Scaling, bias-aware prompting, and test-time reasoning. For all experiments, we kept the base setup from RQ1.

\bmhead{Scaling experiment}
To investigate the effect of scaling model parameters on anatomical bias, we utilized the Qwen3-VL model family because it represents the current state-of-the-art among open-weight vision-language models and provides a consistent architecture across a wide range of model sizes, including the 2B, 4B, 8B, 30B, and  235B variants. The two largest variants (30B and 235B) are mixture-of-experts (MoE) architectures, and the reported parameter counts denote the total number of parameters in each model.

\bmhead{Prompting experiment}
To evaluate whether alternative prompting strategies could mitigate anatomical bias, while otherwise maintaining the same experimental conditions as in RQ1, we tested two additional prompt variants:

\begin{itemize}
    \item "Rarity-aware": "You are a medical expert. Answer the following question solely based on the attached medical image, keeping in mind that it might contain rare cases or non-standard anatomy: {question}."
    \item “Medical expert": "You are a medical expert, answer the following question solely based on the attached medical image, ignoring population biases: {question}."
\end{itemize}
We conducted this experiment on eight state-of-the-art VLMs spanning both closed-weight and open-weight systems: GPT-5, Gemini 2.5 Pro, Llama 4 Maverick, Grok 4, Mistral Medium 3.1, Claude 4.5 Sonnet, Qwen3-VL 235B, and GLM-4.5V.

In a separate experiment, we investigated whether allowing models to express uncertainty would reduce erroneous responses on atypical cases. To this end, we repeated the RQ1 experiment with an additional response option by appending the following suffix to the prompt: "However, if you are unsure, you may answer 'Unsure'."

\bmhead{Reasoning experiment}
To investigate whether extended reasoning improves performance, we evaluated three frontier models that allowed for setting specific reasoning budgets: GPT-5, Claude Sonnet 4.5, and Gemini 2.5 Pro. Each model was tested at three levels of reasoning effort, designated as "low," "medium," and "high":
\begin{itemize}
    \item For GPT-5, we used the model's native reasoning presets corresponding to these three levels.
    \item For Gemini 2.5 Pro, we configured the maximum number of reasoning tokens to 1,024, 8,192, and 24,576 for the low, medium, and high settings, respectively, following the specifications outlined in the model documentation \cite{geminidocs}.
    \item For Claude Sonnet 4.5, for fairness of the comparison, we matched the reasoning token counts of Gemini 2.5 Pro.
\end{itemize}
Unless otherwise specified, we report the results for the “medium” reasoning preset.
\bmhead{Bounding box experiment}

To investigate whether the observed failures primarily originate from language-level priors, we conducted an additional experiment using grounded object detection. Specifically, we prompted Qwen3-VL 235B and Gemini 2.5 Pro, both of which support bounding box generation \cite{2511.21631,2507.06261}, to explicitly detect anatomical structures and organs relevant to answering the \textit{AdversarialAnatomyBench} questions. As bounding boxes naturally provide instance counts, we restricted the analysis to the counting questions from \textit{AdversarialAnatomyBench}, for which answers could be derived directly from the number of generated boxes.

Bounding boxes were generated using the following prompt:\\
\textit{"Detect every instance that belongs to the following categories: \{entity\}. The box\_2d should be [ymin, xmin, ymax, xmax] normalized to 0-1000"}

We report overall counting accuracy based on the number of predicted bounding boxes, as well as accuracy on typical images, accuracy on atypical images, bias rate on atypical images, and agreement with model answers from the “Base prompt” experiment.

\subsection*{Data availability}
\textit{AdversarialAnatomyBench} is publicly available on Hugging Face at \url{https://huggingface.co/datasets/IMSY-DKFZ/AdversarialAnatomyBench}. It provides for every image a link to its original source and its licence, together with the benchmark question and labels.

\subsection*{Code availability}
{\tolerance=2000\emergencystretch=3em The code to run the benchmark is available at \url{https://github.com/IMSY-DKFZ/adversarialanatomybench}. The code is written in Python 3.12 and uses PyTorch 2.8.0, Transformers 4.57.1, pandas and Matplotlib; exact versions of all packages are listed in the repository's requirements file.\par}

\backmatter

\bmhead{Acknowledgements}

We thank Dr. Paul Hellerhoff (Center for Radiology and Nuclear Medicine, Klinikum Dritter Orden, Munich, Germany) and Dr. Abid Irshad (Department of Radiology and Radiological Science, Medical University of South Carolina, Charleston, SC, USA) for allowing us to use their horseshoe kidney images and Lukas Petersson for helpful feedback. Permission was sought and obtained from Radiopaedia.org (ID:202510-0004) prior to research being undertaken. Radiopaedia.org content is used under a CC BY-NC-SA licence. The following Radiopaedia.org images were included in our figures \cite{Jones2013,Uribe2016,Singh2010,Bickle2015,Schubert2011,Gaillard2015,Hacking2015,DiMuzio2012,SantizoCastillo2022,Dixon2015,Qureshi2018,AlSalam2009,Pai2014,Niknejad2020,Knipe2015,Lukies2023,Khurfan2021,Liew2023,Elthokapy2021,Ranchod2023}. Fig. \ref{fig7} and Supplementary Fig.~\ref{fig_b8_aug_exp} additionally include Radiopaedia.org images from \cite{Qaqish2020,AlSalam2010,English2023,Schubert2012}. The kidney image in Fig. \ref{fig1} was provided by Dr. Irshad.
We thank Angela Linan Ebersbach, Linus Schott and Franz Wetzel for their support in data annotation, Dominik Michael for his help in the setup of the annotation infrastructure, and Patrick Godau and Alex Seitel for their support in publishing the code and data.

\bmhead{Funding}
P.K. discloses support for the research of this work from the Helmholtz Association under the joint research school “HIDSS4Health – Helmholtz Information and Data Science School for Health”. T. R. was supported by a scholarship from the Hanns Seidel Foundation with funds from the Federal Ministry of Education and Research Germany (BMBF). F.R.K. receives support from the German Cancer Research Center (CoBot 2.0), the Joachim Herz Foundation (Add-On Fellowship for Interdisciplinary Life Science), the Central Indiana Corporate Partnership AnalytiXIN Initiative, the Evan and Sue Ann Werling Pancreatic Cancer Research Fund, and the Indiana Clinical and Translational Sciences Institute (EPAR4157) funded, in part, by Grant Number UM1TR004402 from the National Institutes of Health, National Center for Advancing Translational Sciences, Clinical and Translational Sciences Award. The content is solely the responsibility of the authors and does not necessarily represent the official views of the National Institutes of Health. A.R. discloses support for the research of this work from the Helmholtz Association of German Research Centers in the scope of the Helmholtz Imaging Incubator (HI). This work is supported by the German Federal Ministry of Education and Research (DECIPHER-M, 01KD2420F). The funders had no role in study design, data collection and analysis, decision to publish or preparation of the manuscript.

\bmhead{Author contributions}
F.R.K., L.M., L.M.-H. and P.K. conceived and designed the study. C.E. and F.R.K. identified atypical conditions and proposed question formulations. C.E., F.R.K., L.M., L.M.-H. and P.K. designed and performed the data collection. C.E. and F.R.K. conducted quality control. C.E., F.R.K., L.M., L.M.-H., M.K., P.K. and T.R. developed the ablation experiments. L.M. and P.K. analysed the data. A.R. and E.C. provided statistical guidance and reviewed the implementation and results. A.R., L.M., L.M.-H. and P.K. wrote the manuscript. A.R., C.E., E.C., F.R.K., L.M., L.M.-H., M.K., P.K. and T.R. revised the manuscript. L.M. and P.K. contributed equally and each may list themselves as lead author on their CV. All authors read and approved the final manuscript.

\section*{Competing interests}
F.R.K. declares an active advisory role for Scopia AI, Canada, unrelated to the research presented here. The remaining authors declare no competing interests.

\bibliography{bibliography}
\newpage

\phantomsection
\section*{Extended Data}

\renewcommand{\figurename}{Extended Data Fig.}
\renewcommand{\tablename}{Extended Data Table}
\renewcommand{\thefigure}{\arabic{figure}}
\renewcommand{\thetable}{\arabic{table}}
\renewcommand{\theHfigure}{ED\arabic{figure}}
\renewcommand{\theHtable}{ED\arabic{table}}
\setcounter{figure}{0}
\setcounter{table}{0}

\begin{figure}[!htbp]
\centering
\includegraphics[width=0.9\textwidth]{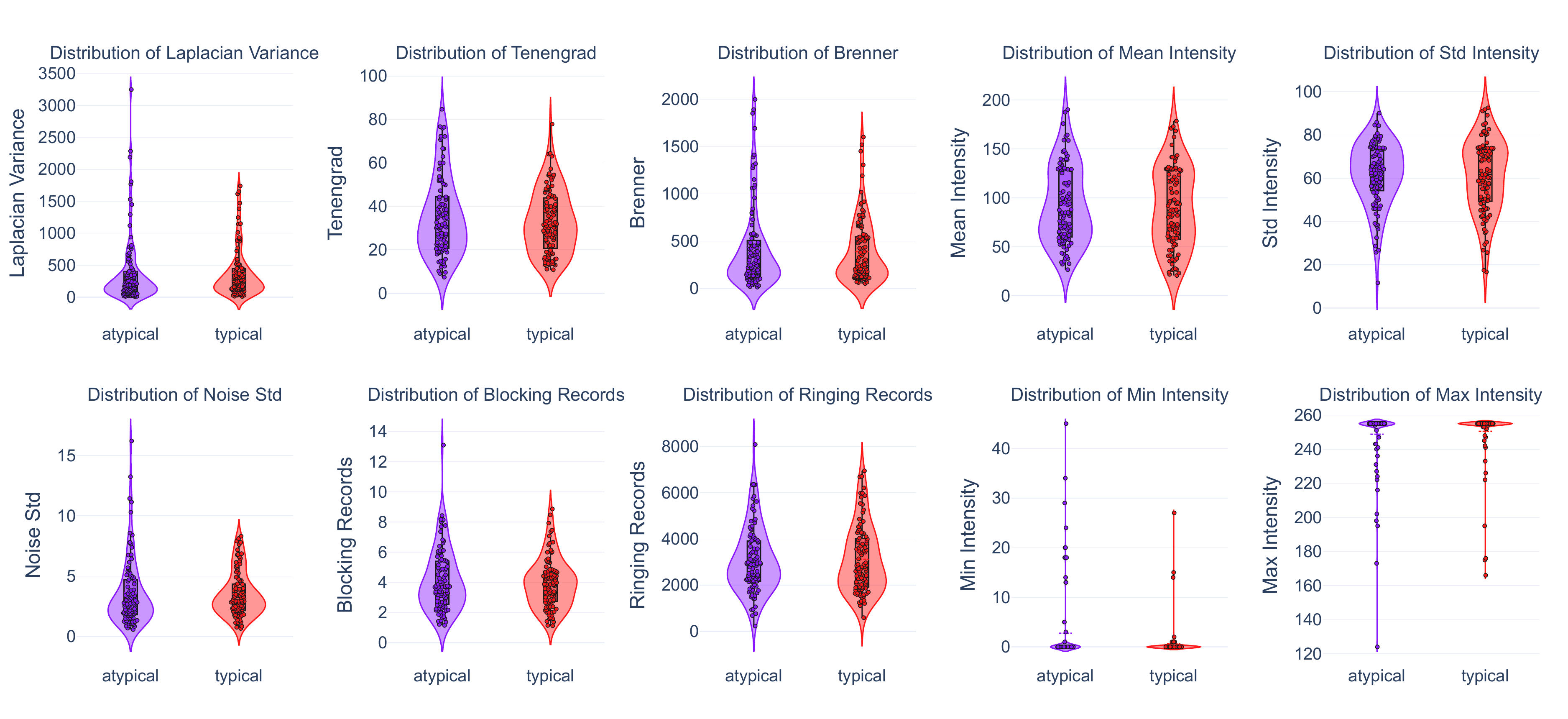}
\caption{\textbf{Image quality metrics indicate comparable quality between atypical and typical images in \textit{AdversarialAnatomyBench}.} Violin plots show the distributions of image-quality metrics computed using Pixel Patrol \cite{pixelpatrol}, stratified by atypical and typical images. Violin widths represent Gaussian kernel-density estimates; boxes span the interquartile range, solid lines indicate medians, dashed lines indicate means, whiskers extend to the most extreme values within 1.5$\times$ the interquartile range, and points represent individual images. n = 100 typical and n = 100 atypical images.
}\label{supfig_c2_imagequalitymetrics}
\end{figure}

\begin{figure}[!htbp]
\centering
\includegraphics[width=0.9\textwidth]{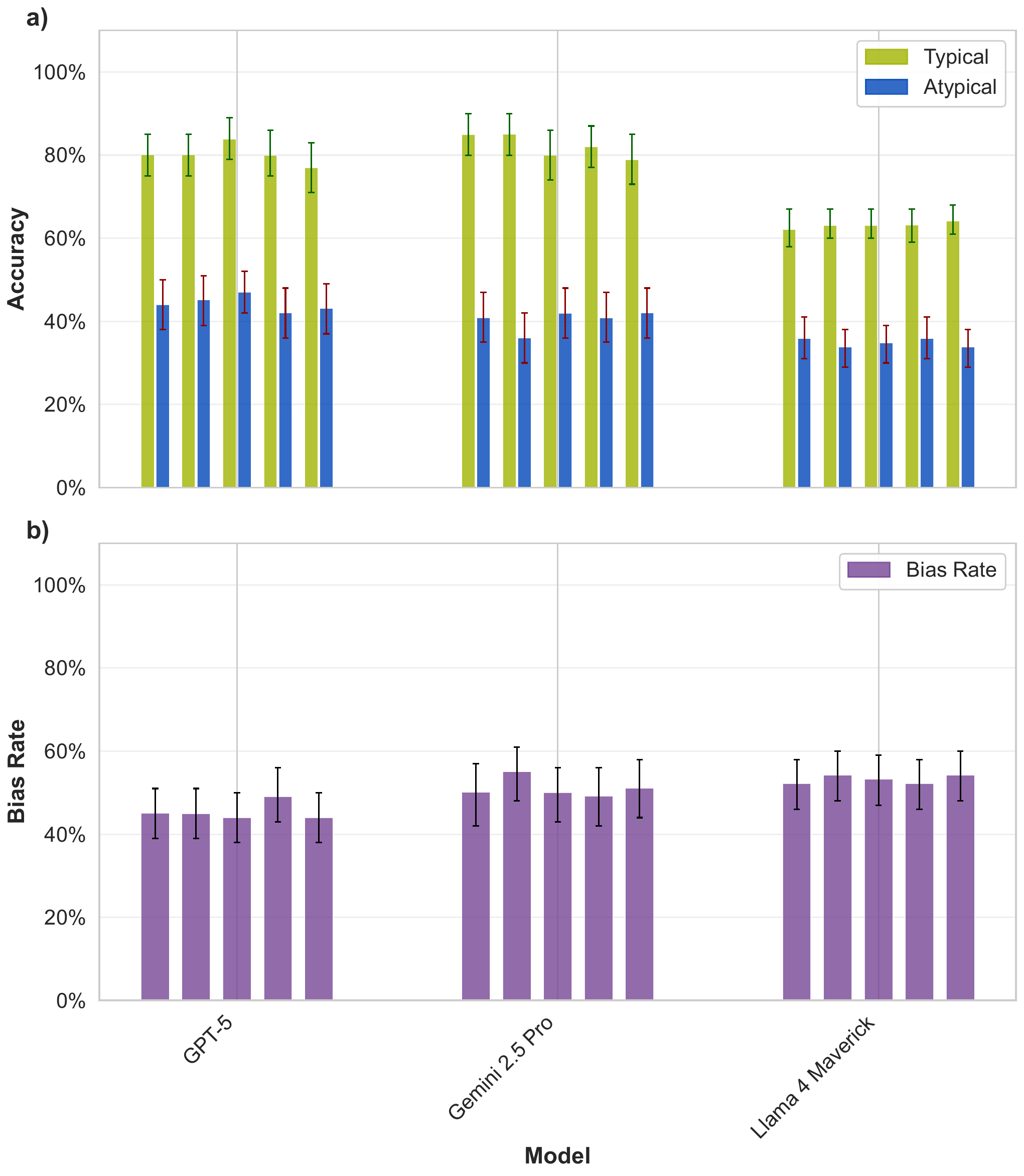}
\caption{\textbf{The performance gap between typical and atypical images
persists consistently across five independent runs.} The benchmark was repeated
five times in the base configuration, using identical inference settings, for
GPT-5 (medium reasoning), Gemini 2.5 Pro (medium reasoning) and Llama 4
Maverick. Each pair of bars shows accuracy on typical and atypical images for a
single run. Bars show the mean and error bars the 95\% confidence interval (2.5th--97.5th percentile) of 1,000 stratified bootstrap resamples, each on n = 100 typical and n = 100 atypical images from 34 questions per run. (a) Accuracy. (b) Bias rate.}\label{fig_b6_variance}
\end{figure}

\begin{figure}[!htbp]
\centering
\includegraphics[width=0.9\textwidth]{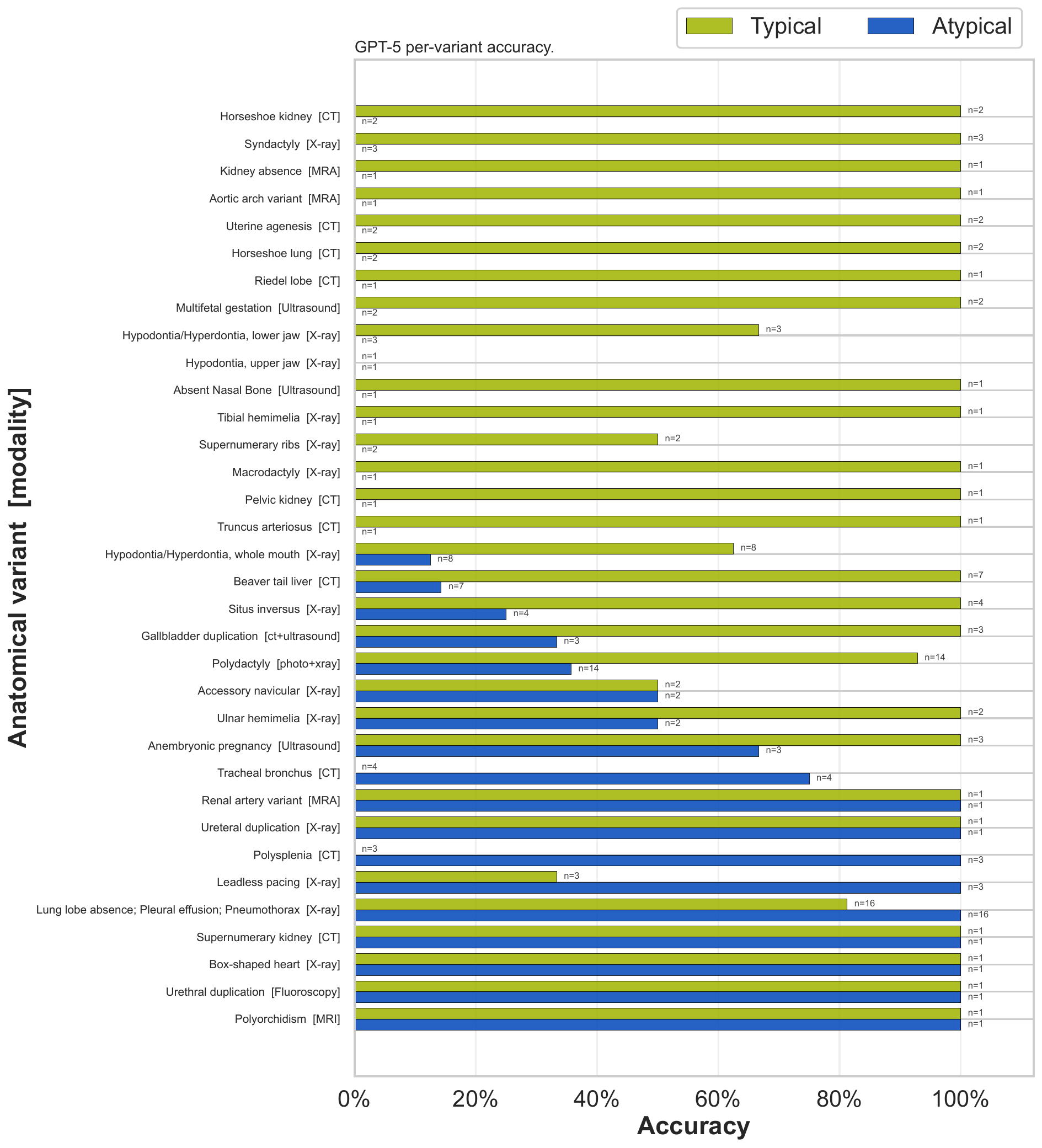}
\caption{\textbf{Per-variant analysis highlights limited performance of GPT-5 across atypical image variants.} Accuracy on typical (green) and atypical (blue) images grouped by question type. The condition of the atypical images, as well as the modalities and number of images contributing to each estimate, is given on the left side. GPT-5 scores 0\% on 16 of 34 variants, highlighting the performance difference. Weighting each variant equally instead of each image widens GPT-5's accuracy gap between typical and atypical anatomy from 36.0 to 46.3 pp. Higher accuracy is preferred.}\label{fig_a2_variant_analysis}
\end{figure}

\begin{figure}[!htbp]
\centering
\includegraphics[width=0.9\textwidth]{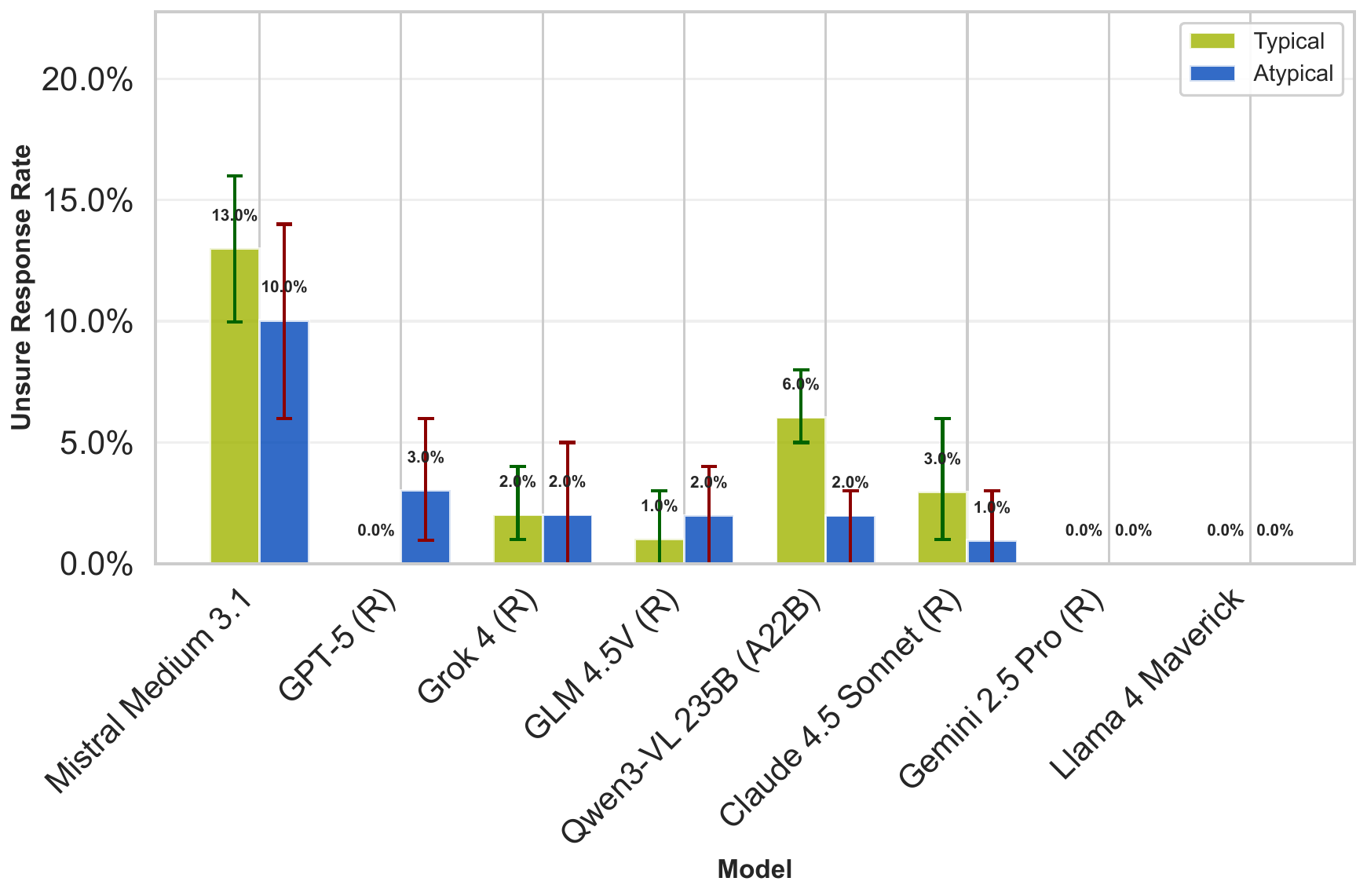}
\caption{\textbf{Models rarely choose to answer with “Unsure” when provided with a possibility to do so.} The unsure answer is chosen only in up to 13\% of typical (green) and 10\% atypical (blue) cases, with some models, including Gemini 2.5 Pro and Llama 4 Maverick never providing unsure responses. Bars show the mean and error bars the 95\% confidence interval (2.5th--97.5th percentile) of 1,000 stratified bootstrap resamples, each on n = 100 typical and n = 100 atypical images from 34 questions. Reasoning models are marked with (R).}\label{supfig_unsure}
\end{figure}

\begin{figure}[!htbp]
\centering
\includegraphics[width=0.9\textwidth]{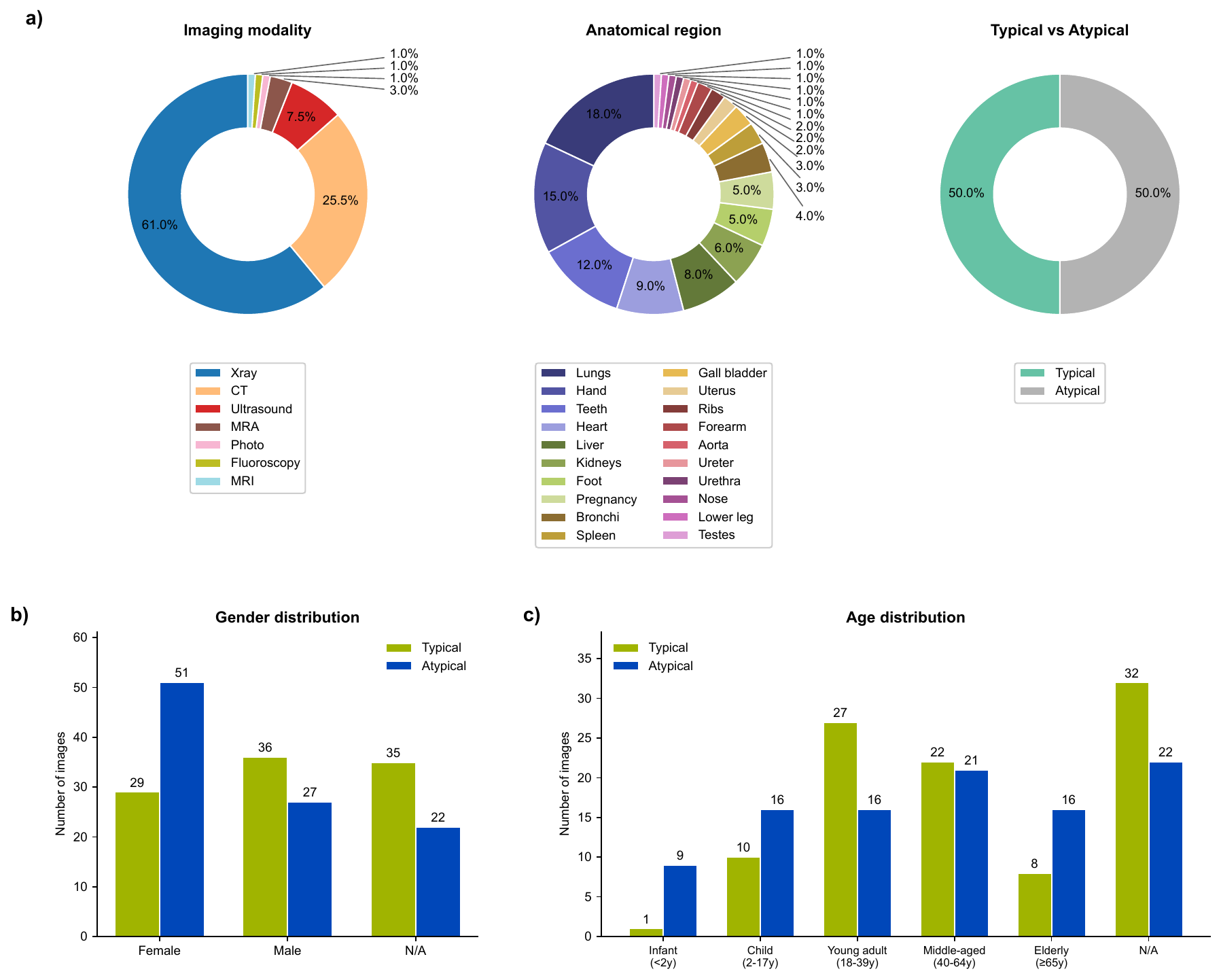}
\caption{\textbf{AdversarialAnatomyBench contains 200 images covering 20 anatomical regions and seven imaging modalities, with a balanced number of typical and atypical cases (a).} Gender (b) and age (c) distributions for typical and atypical images extracted from CheXpert and Radiopaedia case metadata where available (n=143 with gender, n=146 with age, out of 200 images). Images sourced from other datasets do not include demographic metadata.
}\label{supfig_c1_datasetoverview}
\end{figure}

\begin{table}[!htbp]
\centering\small
\begin{tabular}{@{}lllll@{}}
\toprule
Model & $\uparrow$ Accuracy (atypical) & $\uparrow$ Accuracy (typical) & $\downarrow$ Bias rate \\
\midrule
GPT-5 (high) (R) & \meanCIpar{43.0}{38.0}{49.0} &\meanCIpar{84.9}{79.0}{90.0} &\meanCIpar{45.0}{38.0}{51.0} \\
GPT-5 (medium) (R) & \meanCIpar{43.9}{38.0}{50.0} &\meanCIpar{79.9}{74.0}{86.0} &\meanCIpar{49.1}{42.0}{55.0} \\
GPT-5 (low) (R) & \meanCIpar{42.9}{37.0}{49.0} &\meanCIpar{78.9}{73.0}{85.0} &\meanCIpar{48.0}{41.0}{54.0} \\
\midrule
Gemini 2.5 Pro (high) (R) & \meanCIpar{44.0}{37.0}{50.0} &\meanCIpar{83.0}{77.0}{89.0} &\meanCIpar{51.0}{44.0}{58.0} \\
Gemini 2.5 Pro (medium) (R) & \meanCIpar{40.0}{34.0}{46.0} &\meanCIpar{81.0}{76.0}{86.0} &\meanCIpar{50.0}{43.0}{57.0} \\
Gemini 2.5 Pro (low) (R) & \meanCIpar{42.9}{38.0}{49.0} &\meanCIpar{82.9}{78.0}{88.0} &\meanCIpar{52.1}{45.0}{58.0} \\
\midrule
Claude 4.5 Sonnet (high) (R) & \meanCIpar{27.0}{23.0}{31.0} &\meanCIpar{78.9}{73.0}{84.0} &\meanCIpar{69.1}{64.0}{74.0} \\
Claude 4.5 Sonnet (medium) (R) & \meanCIpar{26.0}{21.0}{31.0} &\meanCIpar{81.1}{75.0}{86.0} &\meanCIpar{69.0}{64.0}{75.0} \\
Claude 4.5 Sonnet (low) (R) & \meanCIpar{28.9}{25.0}{32.0} &\meanCIpar{76.1}{70.0}{82.0} &\meanCIpar{66.1}{62.0}{70.0} \\
\botrule
\end{tabular}
\caption{\textbf{Reasoning experiment results.} Increased amount of reasoning does not result in improved performance on atypical cases. Values are the mean and, in parentheses, the 95\% confidence interval (2.5th--97.5th percentile) of 1,000 stratified bootstrap resamples, each on n = 100 typical and n = 100 atypical images from 34 questions per model and reasoning effort. Reasoning models are denoted with (R).}\label{suptab2}%
\end{table}

\begin{table}[!htbp]
\centering\small
\begin{tabular}{@{}llllll@{}}
\toprule
Model & $\uparrow$ Accuracy & \makecell{$\uparrow$ Accuracy\\ (atypical)} & \makecell{$\uparrow$ Accuracy\\ (typical)} & $\downarrow$ Bias rate & \makecell{Agreement with\\``Base prompt''\\experiment}\\
\midrule
Gemini 2.5 Pro & \meanCIstacked{46.7}{37.7}{55.7} &\meanCIstacked{9.9}{4.9}{14.8}&\meanCIstacked{83.7}{77.0}{90.2} & \meanCIstacked{73.6}{65.6}{82.0} & \makecell{62.0\%} \\
Qwen3-VL 235B & \meanCIstacked{37.8}{31.1}{44.3} &\meanCIstacked{18.0}{11.5}{24.6} &\meanCIstacked{57.5}{52.5}{62.3} & \meanCIstacked{49.3}{42.6}{55.7} & \makecell{46.3\%} \\
\botrule
\end{tabular}
\caption{\textbf{When generating bounding boxes the models still exhibit a large performance gap between atypical and typical images on counting questions.} Values are the mean and, in parentheses, the 95\% confidence interval (2.5th--97.5th percentile) of 1,000 stratified bootstrap resamples, each on the n = 61 typical and n = 61 atypical images of the 23 counting questions; agreement with the ``Base prompt'' experiment is given without a confidence interval.}\label{tab_a3_bb}
\end{table}

\begin{table}[!htbp]
\centering\small
\begin{tabular}{@{}lllll@{}}
\toprule
Organization & Name  & Medical & Access & Model Size\\
\midrule
Alibaba  & Qwen3-VL 235B \cite{2511.21631} & No  & Open & 235B (A22B)\\
   & Qwen3-VL 30B \cite{2511.21631} & No  & Open & 30B (A3B)\\
   & Qwen3-VL 8B \cite{2511.21631} & No  & Open & 8B\\
   & Qwen3-VL 4B \cite{2511.21631} & No  & Open & 4B\\
   & Qwen3-VL 2B \cite{2511.21631} & No  & Open & 2B\\
   & Lingshu-7B  \cite{lingshu} & Yes & Open & 7B \\
Anthropic  & Claude 4.5 Sonnet (R) \cite{anthropic2025sonnet45card} & No  & Closed & -\\
   & Claude 4.5 Haiku (R) \cite{anthropic2025haiku45card} & No  & Closed & -\\
Google   & Gemini 2.5 Pro (R) \cite{2507.06261}  & No & Closed & - \\
   & Gemini 2.5 Flash (R) \cite{2507.06261} & No & Closed & - \\
   & Gemini 2.5 Flash Lite (R) \cite{2507.06261} & No & Closed & - \\
   & Gemma 3 27B \cite{gemmateam2025gemma3technicalreport}   & No & Open & 27B \\
   & Gemma 3 4B \cite{gemmateam2025gemma3technicalreport}   & No & Open & 4B \\
   & MedGemma 4B \cite{sellergren2026medgemmatechnicalreport}  & Yes & Open & 4B \\
MBZUAI & MedMO-4B \cite{deria2026medmogroundingunderstandingmultimodal} & Yes & Open & 4B \\
   & MedMO-8B Next \cite{deria2026medmogroundingunderstandingmultimodal} & Yes & Open & 8B \\
Meta   & Llama 4 Maverick \cite{metaai2025llama4} & No & Open & 400B (A17B)\\
   & Llama 4 Scout \cite{metaai2025llama4} & No & Open & 109B (A17B)\\
Mistral  & Mistral Medium 3.1 \cite{mistral2025medium31} & No  & Closed & -\\
  & Mistral Small 3.2 24B \cite{mistral2025small32} & No  & Open & 24B\\
OpenAI    & GPT-5 (R) \cite{gpt5_card}  & No & Closed & - \\
   & GPT-5 mini (R) \cite{gpt5_card} & No & Closed & - \\
   & GPT-5 nano (R) \cite{gpt5_card} & No & Closed & - \\
xAI   & Grok 4 (R) \cite{xai2025grok4} & No & Closed & -\\
zAI  & GLM 4.5V (R) \cite{vteam2026glm45vglm41vthinkingversatilemultimodal} & No  & Open & 106B (A12B)\\
\botrule
\end{tabular}
\caption{\textbf{Model Overview.} Our evaluation included 25 models from leading model families and included both open and closed-weight models of varying sizes. Reasoning models are denoted with (R) and the numbers in parentheses in the “Model Size” column provide the active parameter count for case of \acf{MoE} models.}\label{suptab1}
\end{table}

\clearpage
\section{Supplementary Information}\label{secSI}

\renewcommand{\figurename}{Supplementary Fig.}
\renewcommand{\tablename}{Supplementary Table}
\renewcommand{\thefigure}{\arabic{figure}}
\renewcommand{\thetable}{\arabic{table}}
\renewcommand{\theHfigure}{S\arabic{figure}}
\renewcommand{\theHtable}{S\arabic{table}}
\setcounter{figure}{0}
\setcounter{table}{0}

\begin{figure}[!htbp]
\centering
\includegraphics[width=0.9\textwidth]{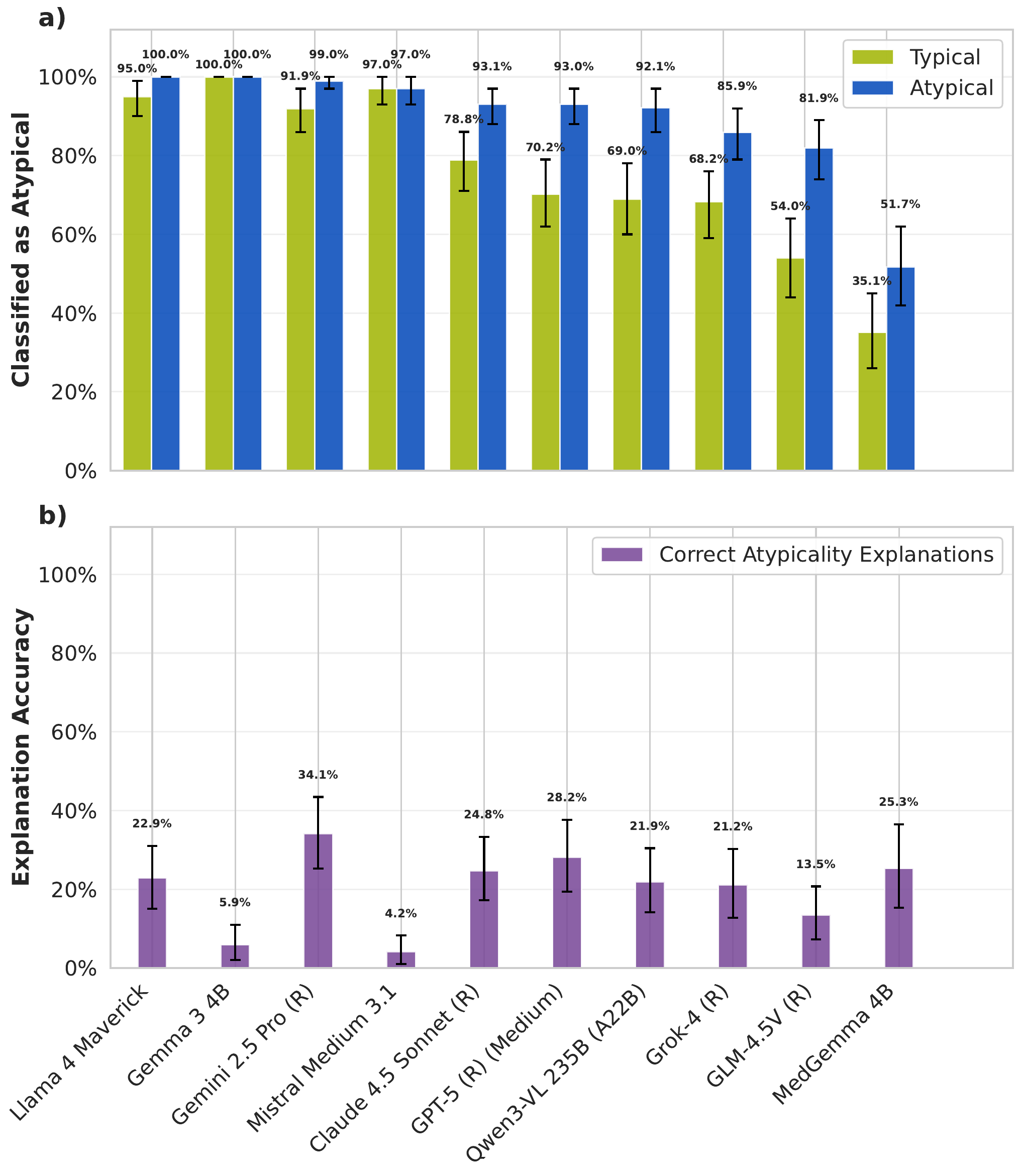}
\caption{\textbf{Most tested models classify the majority of typical images in \textit{AdversarialAnatomyBench} as atypical (a) and do not provide the underlying medical condition as the justification for their answer (b).} Explanation accuracy was computed for True Positives (atypical images classified as such). The correctness of the explanations was judged in relation to the atypicalities considered in the main benchmark questions. In principle, there could be other atypicalities present. (a) n = 100 typical and n = 100 atypical images per model. (b) n = 100 (Llama 4 Maverick), 100 (Gemma 3 4B), 99 (Gemini 2.5 Pro), 97 (Mistral Medium 3.1), 93 (Claude 4.5 Sonnet), 93 (GPT-5), 92 (Qwen3-VL 235B), 86 (Grok 4), 82 (GLM-4.5V) and 52 (MedGemma 4B) true positives. Each model answered each image once. Bars show the mean and error bars the 95\% confidence interval (2.5th--97.5th percentile) of 1,000 bootstrap resamples of the images. (R) highlights reasoning models.
}\label{fig_b1_atypical_class}
\end{figure}

\subsection{Foundational baseline}
To contextualise the difficulty of the benchmark, we introduced an additional baseline task where the same question was asked about every image, attempting to classify it as typical or atypical: \textit{"Can you see anything atypical in the attached medical image? Answer with either \textbackslash"Yes\textbackslash" or \textbackslash"No\textbackslash". For the cases where you answer with \textbackslash"Yes\textbackslash" provide a justification. Format your answer as Answer: Yes/No \textbackslash{}n Justification: \textless{}text\textgreater{}.”} We assessed both the models’ answers as well as the justifications they provide to classify the images as atypical. For this task, images were paired with respect to the anatomical structure targeted by the benchmark question and might still contain unrelated findings.

We assessed answer justifications using a large-language-model-as-a-judge framework with GPT-5 and the following prompt: \\
“\textit{You are an expert medical question answering judge. \\
Failure mode concept: \{failure\_mode\} \\
Ground truth: the failure mode is present in this image. \\
The vision-language model previously answered: \{short\_answer\}. \\
Model explanation: \\
\hspace*{1em}\{model\_explanation\} \\
Decide if the explanation correctly describes the failure mode concept. \\
Respond with a single word: Correct or Incorrect. \\
Answer: }“ \\
Where the failure mode is the underlying rare variant.

We report the results for the models included in Fig. 3 in the main experiment, namely: GPT-5 (medium), Gemini 2.5 Pro (medium), Llama 4 Maverick, Grok 4, Claude 4.5 Sonnet (medium), Qwen3-VL 235B, GLM-4.5V, Mistral Medium 3.1, Gemma 3 4B, MedGemma 4B.

\subsection{Prompt ablations}
To further examine the sensitivity of benchmark performance to prompt format, we evaluated nine additional prompting strategies. These strategies varied three prompt characteristics: \\
\textbf{Phrasing}. We modified prompt wording while preserving the original intent and answer format:
\begin{itemize}
\item \textbf{Rephrasing 1 (polite, passive)}: "Based on the provided medical image, please answer the following question: "
\item \textbf{Rephrasing 2 (direct, imperative)}: "Examine the medical image and respond to this question: "
\item \textbf{Rephrasing 3 (terse)}: "Using only the medical image below, answer: "
\end{itemize}
\textbf{Structure.} These prompts varied in how models were instructed to organize their responses:
\begin{itemize}
\item \textbf{Chain-of-thought}: "Think step by step before giving your final answer."
\item \textbf{Answer first}: "First state your answer, then explain your reasoning."
\item \textbf{Reflection}: "After arriving at your initial answer, reflect on whether it could be wrong and reconsider."
\end{itemize}
\textbf{Constraints.} These prompts restricted the basis or process of reasoning:
\begin{itemize}
\item \textbf{Visual only}: "Only take into account what is actually visible in the image. Do not rely on prior knowledge, assumptions, or population-level statistics."
\item \textbf{Describe first}: "First, describe in detail what you observe in the image. Then, based solely on your description, answer the question."
\item \textbf{German thinking}: "Reason about the image entirely in German. Then, provide your final answer in English." This prompt was intended to test whether switching the language of the internal reasoning allows the model to bypass the English-trained priors.
\end{itemize}
These additional prompting strategies were evaluated for GPT-5, Gemini 2.5 Pro, and Llama 4 Maverick.

\begin{figure}[!htbp]
\centering
\includegraphics[width=0.9\textwidth]{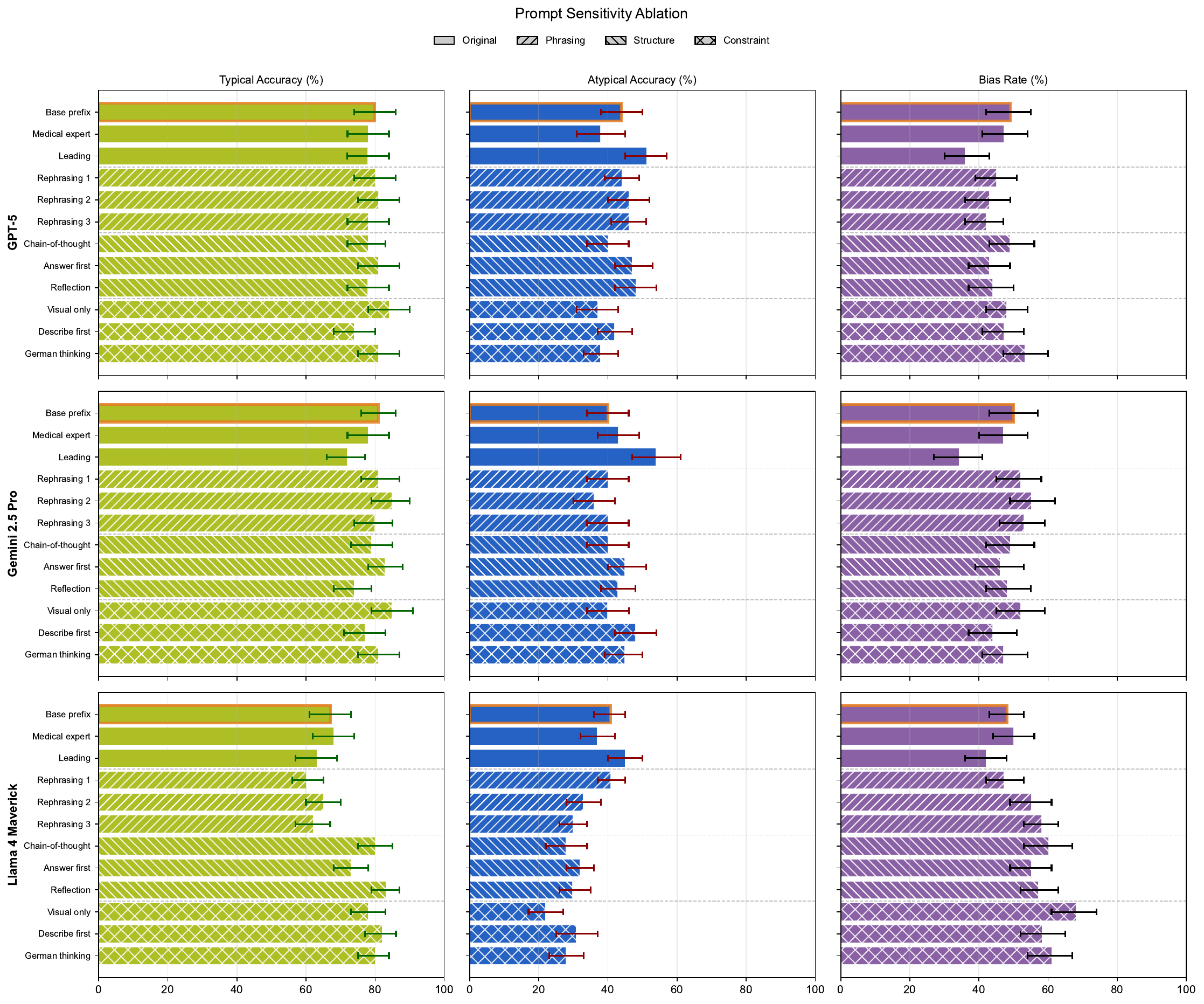}
\caption{\textbf{Varying prompt phrasing, structure, or constraints did not lead to closing the performance gap between typical (green) and atypical (blue) images.} Rows show GPT-5, Gemini 2.5 Pro and Llama 4 Maverick; columns show accuracy on typical (green) and atypical (blue) images and the bias rate (purple) for the three prompts of the main experiments (Base prefix, Medical expert, Leading) and the nine ablations. Hatching indicates the prompt category; the base prompt is outlined in orange. Bars show the mean and error bars the 95\% confidence interval (2.5th--97.5th percentile) of 1,000 stratified bootstrap resamples, each on n = 100 typical and n = 100 atypical images from 34 questions; each model answered each image once per prompt.
}\label{fig_b2_prompt_abl}
\end{figure}

\subsection{In-context examples}
In addition to scaling model size, explicit prompting about possible rare anatomical variants, and allocating additional test time compute explored in the main paper, we evaluated in-context learning (ICL) as an alternative mitigation strategy. We assessed its viability for three models: GPT-5 (medium reasoning), Gemini 2.5 Pro (medium reasoning), and Llama 4 Maverick. Specifically, we selected 170 image-question pairs for which at least one additional atypical image and one typical image of the same question type were available. For each pair, we randomly assigned these images as “Example 1” and “Example 2” in the following prompt:

``{\itshape I will show you two examples, then ask you to answer the same question for a new image.

Example 1: \\
{}[image] \\
Question: \{question\} \\
Correct answer: \{example1\_gt\_answer\} \\

Example 2: \\
{}[image] \\
Question: \{question\} \\
Correct answer: \{example2\_gt\_answer\} \\

Now answer for this image: \\
{}[image] \\
Answer the following question based on the attached medical image: \\\{question\} \{question\_spec\}}``

The ordering of typical and atypical examples was randomized. Alongside the ICL results, we also report results obtained using the Base prompt on the same subset of 170 questions.

\begin{figure}[!htbp]
\centering
\includegraphics[width=0.9\textwidth]{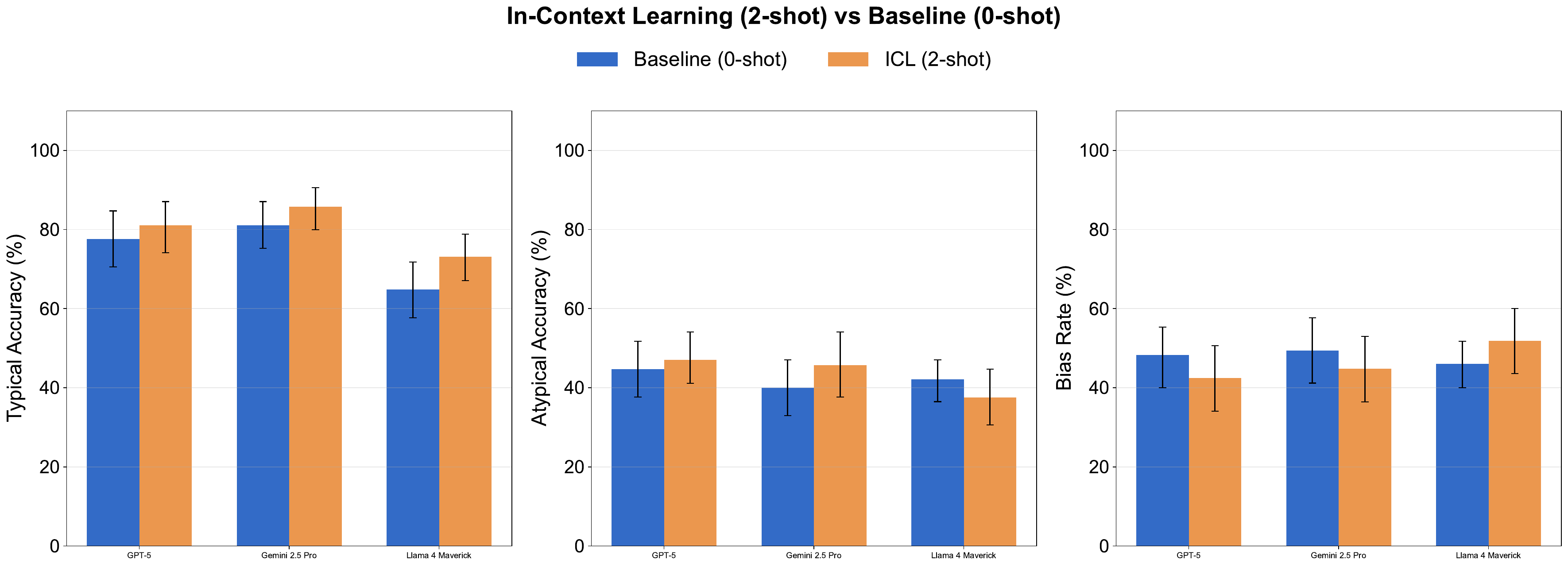}
\caption{\textbf{Providing in-context examples does not close the performance gap between typical and atypical images.} Accuracy on typical (left) and atypical (centre) images and bias rate (right) without (Baseline, 0-shot, blue) and with in-context examples (ICL, 2-shot, orange), i.e. one typical and one atypical example image of the same question, drawn from the other benchmark images. The evaluation uses the n = 85 typical and n = 85 atypical images from the 19 questions with at least two typical and two atypical images, each answered once per condition. Bars show the mean and error bars the 95\% confidence interval (2.5th--97.5th percentile) of 1,000 stratified bootstrap resamples.
}\label{fig_b3_icl}
\end{figure}

\subsection{Vision embeddings} \label{subsec:visionembeddings}
Understanding where errors originate within modern VLM pipelines - whether in the visual encoder, projection layers, decoder, or downstream prompting - is itself an important open research question and beyond the scope of the present work. As an initial exploratory analysis, we investigated whether atypical anatomy is reflected in the global visual embedding space of two open-weight VLMs: Qwen3-VL 8B and MedMO-8B-Next. We selected this pair because both models share a common architecture, while one is trained as a general-purpose VLM and the other is specialized for medical applications. Additionally, we examined how common image augmentations affect the resulting embeddings.

Each image was passed through the models’ vision encoders to obtain patch-wise embeddings, which were subsequently averaged to produce a single 4096-dimensional embedding per image. Prior to embedding extraction, images were resized using the same preprocessing pipeline as in the main experiments. We additionally computed embeddings for augmented versions of each image using the Medium augmentation preset from the augmentation experiment described in the section ``\nameref{subsec:imageaugexp}'' below.

\begin{figure}[!htbp]
\centering
\includegraphics[width=0.9\textwidth]{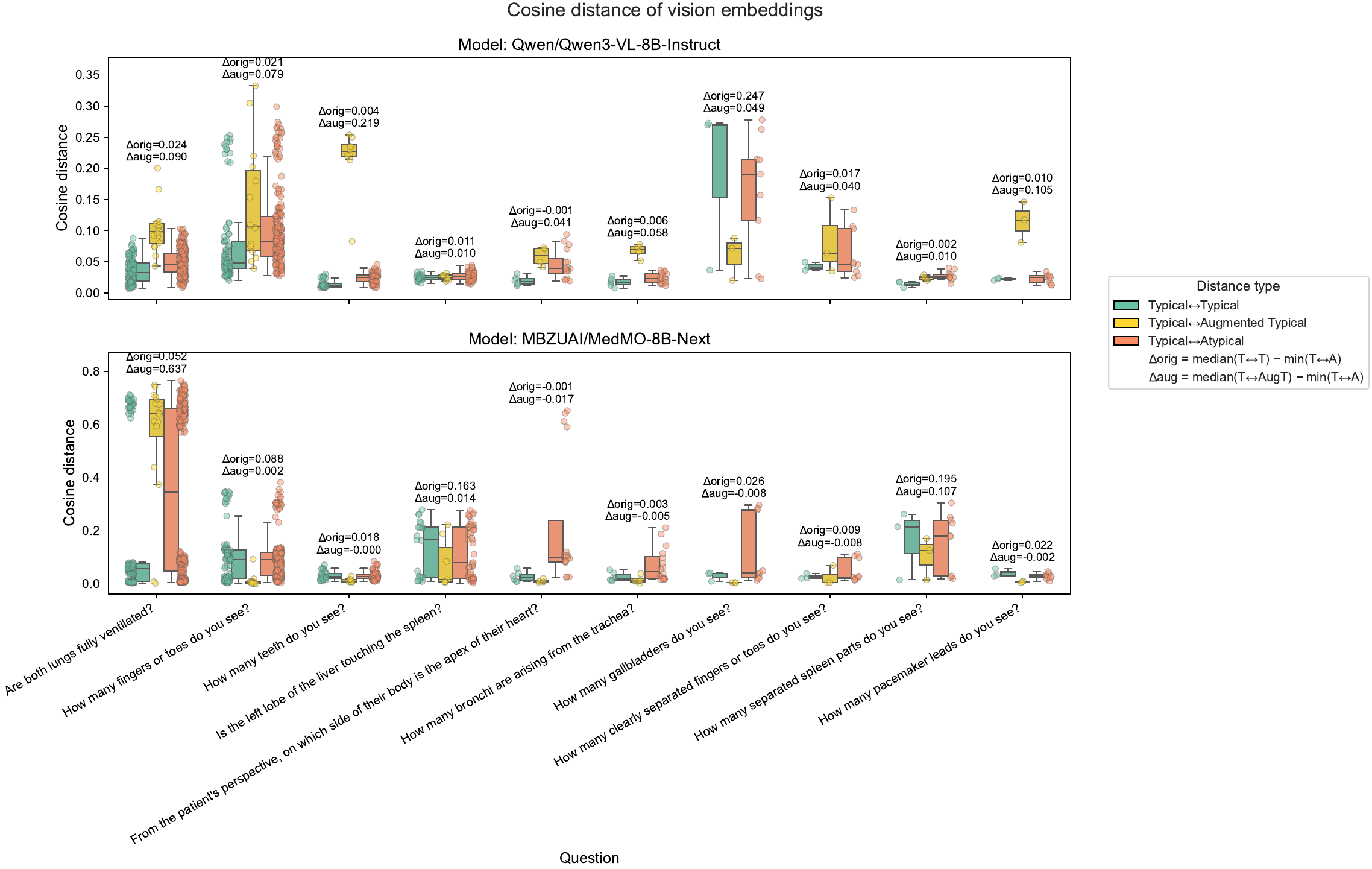}
\caption{\textbf{The closest pair of a typical and an atypical image is usually closer than the median distance between typical images in the vision embedding space of Qwen3-VL 8B and MedMO-8B-Next.} Box plot showing the cosine distance in the embedding space generated by the vision encoders of the two models of images corresponding to the 10 most frequent questions in AdversarialAnatomyBench. Box edges denote the interquartile range (25th–75th percentiles), the horizontal line denotes the median, and whiskers extend to the most extreme points within 1.5× the interquartile range. $\Delta_{\text{orig}}$ denotes the difference between the median pairwise distance among typical images and the minimum typical-atypical distance. $\Delta_{\text{aug}}$ denotes the difference between the median distance for typical-augmented image pairs and the minimum typical-atypical distance. Box colours denote typical--typical pairs (green), each typical image paired with its own augmented copy (yellow) and typical--atypical pairs (orange); every distance is overlaid as a point. n (typical--typical / typical--augmented / typical--atypical distances) per question: 120/16/256 (lungs ventilated), 91/14/196 (fingers or toes), 28/8/64 (teeth), 21/7/49 (liver touching the spleen), 6/4/16 each (apex of the heart; bronchi) and 3/3/9 each for the remaining four questions; identical for both models.
}\label{fig_b4_distance}
\end{figure}

\begin{figure}[!htbp]
\centering
\includegraphics[width=0.9\textwidth]{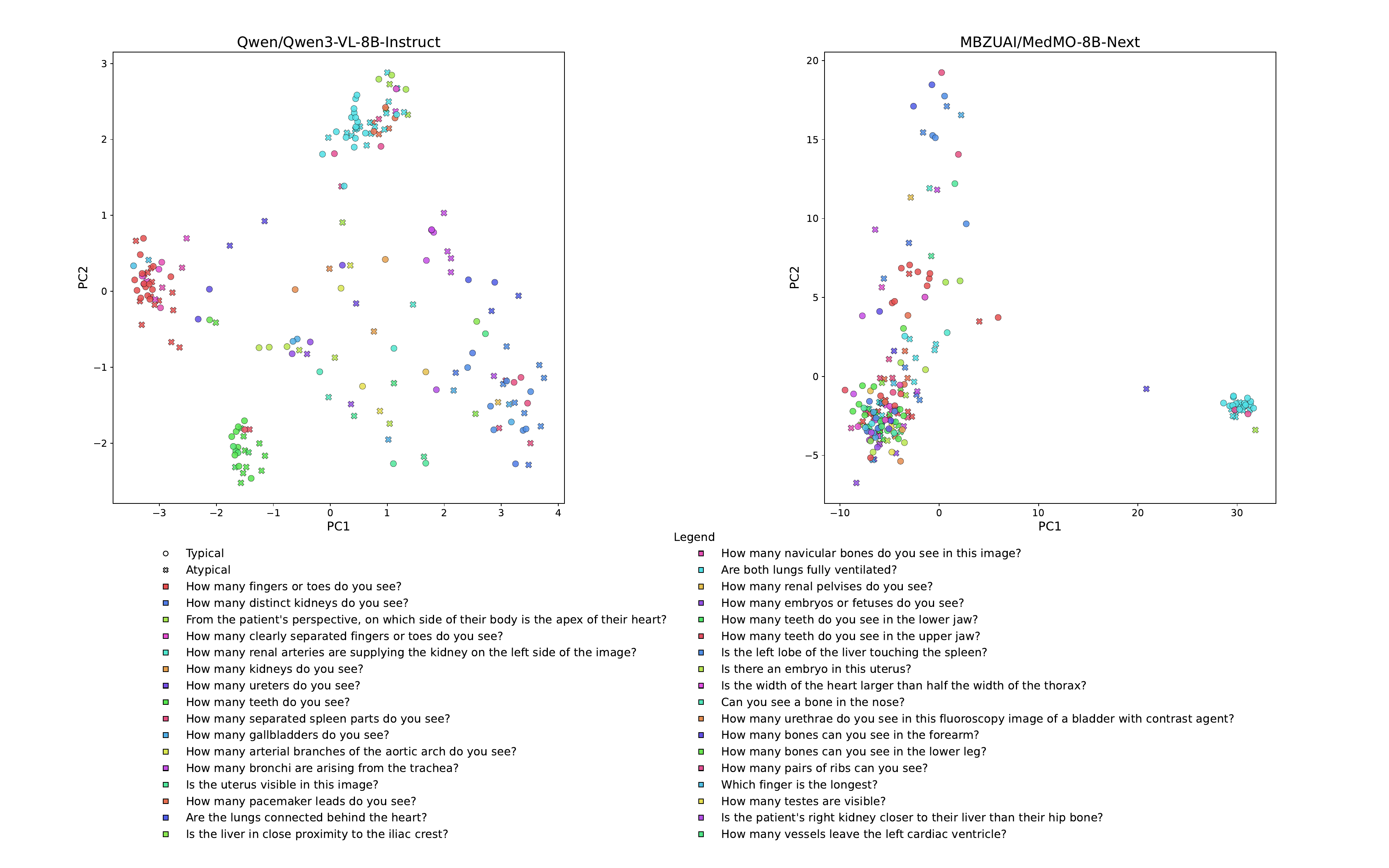}
\caption{\textbf{Vision embeddings from Qwen3-VL 8B and MedMO-8B-Next separate images from questions about different regions of the body.} PCA projection to the first two principal components of vision embeddings produced by Qwen3-VL 8B (left) and MedMO-8B-Next (right). Atypical images (crosses) are usually close to typical images (dots) from the same question. Colours denote the benchmark question. Each panel shows n = 200 images (100 typical and 100 atypical) from all 34 questions; the PCA was fitted separately for each model.
}\label{fig_b5_pca}
\end{figure}

We visualize the first two principal components of the resulting embeddings in scatter plots (Supplementary Fig. \ref{fig_b5_pca}) and report whisker plots of pairwise distances between typical images, between typical and atypical images, and between typical and augmented images (Supplementary Fig. \ref{fig_b4_distance}).
For both models, the median distance between pairs of typical images exceeded the distance between the closest pair of a typical and an atypical image for nine of the ten most frequent questions in \textit{AdversarialAnatomyBench}, using both Euclidean and cosine distance in the pooled embedding space (Supplementary Fig. \ref{fig_b4_distance}). Principal component analysis additionally showed that embeddings cluster primarily according to broad anatomical regions, while typical and atypical images from the same question type often remain close in representation space (Supplementary Fig. \ref{fig_b5_pca}).

Importantly, we do not interpret these analyses as demonstrating that the observed failures originate solely from the vision encoder. Localized atypical structures may be diluted by global pooling, and downstream decoder priors may substantially influence the final prediction. We therefore view these experiments as preliminary evidence that atypical anatomy is not always strongly separated in the global representation space, which may contribute to the observed benchmark failures.

\subsection{Reversing language prior}
\begin{figure}[!htbp]
\centering
\includegraphics[width=0.9\textwidth]{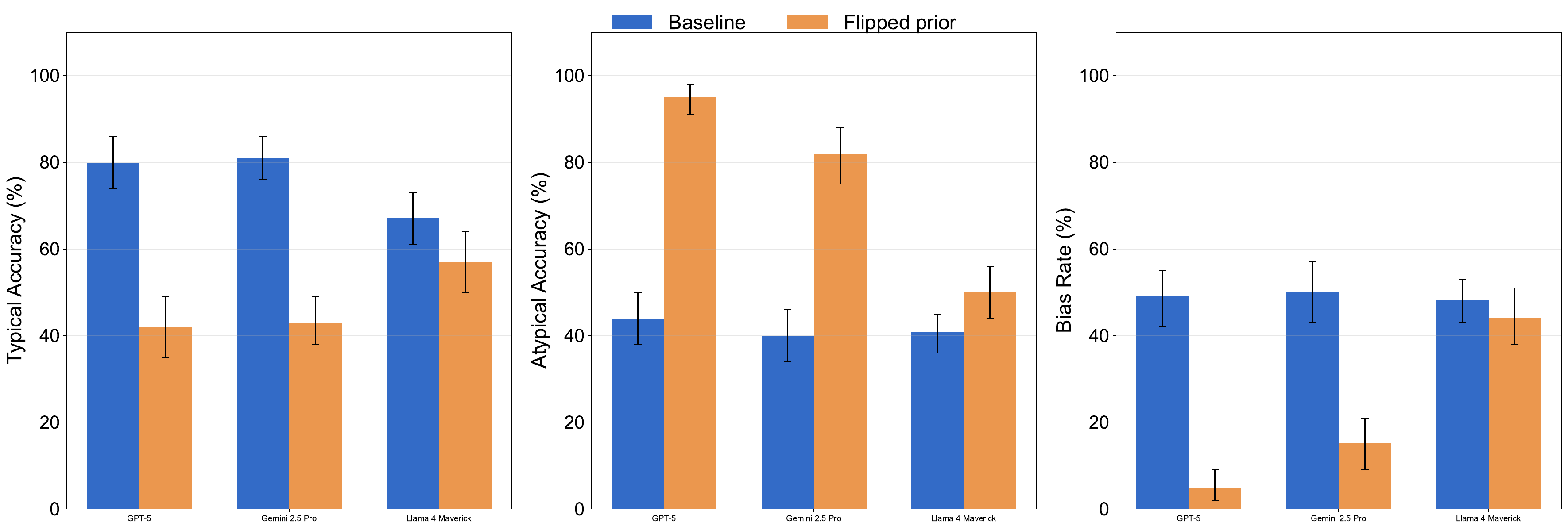}
\caption{\textbf{Reversing the language prior flips the direction of the performance gap, indicating that model predictions are highly sensitive to the prompted prior.} Accuracy on typical and atypical images (left, center) and bias rate (right) under the baseline prompt (blue) and the flipped prior prompt (orange). Bars show the mean and error bars the 95\% confidence interval (2.5th--97.5th percentile) of 1,000 stratified bootstrap resamples, each on n = 100 typical and n = 100 atypical images from 34 questions; each model answered each image once per prompt. Higher accuracy and lower bias rate are preferred.
}\label{fig_b7_reverseprior}
\end{figure}

To directly prompt whether model predictions are driven by language priors or by visual evidence, we conducted an experiment in which the default assumption of typical anatomy was reversed. Specifically, models were told that the image likely showed atypical anatomy and were instructed to answer differently only if direct visual evidence supported an alternative conclusion: \\
\textit{"Answer the following question based on the attached medical image: {question}. The attached image is likely to show atypical anatomy for which the answer to this question is {atypical\_answer}. Answer differently only if the attached image presents direct visual evidence to do so. {question\_spec}"} \\
For atypical images, \textit{{atypical\_answer}} was set to the ground-truth answer. For typical images, it was set to a randomly selected ground-truth answer from atypical images of the same question type, simulating the case where the model is given an incorrect prior. This experiment was performed for GPT-5 (medium reasoning), Gemini 2.5 Pro (medium reasoning), and Llama 4 Maverick. \\
The results (Supplementary Fig. \ref{fig_b7_reverseprior}) show that reversing the prior substantially improves accuracy on atypical images while degrading accuracy on typical images to a comparable level. For GPT-5, atypical accuracy rose from 44\% to 95\% while typical accuracy dropped from 80\% to 42\%. For Gemini 2.5 Pro, atypical accuracy increased from 40\% to 82\% while typical accuracy fell from 81\% to 43\%. For Llama 4 Maverick, the effect was weaker, with atypical accuracy rising from 41\% to 50\% and typical accuracy decreasing from 67\% to 57\%. These findings indicate that model predictions are highly sensitive to the prompted prior: when the prior is flipped, the direction of the performance gap reverses. This experiment establishes sensitivity to prompted priors, but it does not localize where in the pipeline the errors arise, as the flipped instruction could act at any stage from visual attention to answer generation. However, this approach cannot serve as a practical mitigation strategy, as it requires knowledge of the correct answer before querying the model.

\subsection{Image Augmentation Experiment} \label{subsec:imageaugexp}
To investigate whether differences in image quality contribute to the observed performance gap between typical and atypical images, we repeated the base experiment using augmented versions of all benchmark images. We applied three levels of combined image degradations that simultaneously reduced resolution and contrast while adding Gaussian noise: weak (512px resolution, 0.75× contrast, Gaussian noise with $\sigma$=12), medium (256px, 0.5× contrast, $\sigma$=25), and strong (128px, 0.25× contrast, $\sigma$=50).

\begin{figure}[!htbp]
\centering
\includegraphics[width=0.9\textwidth]{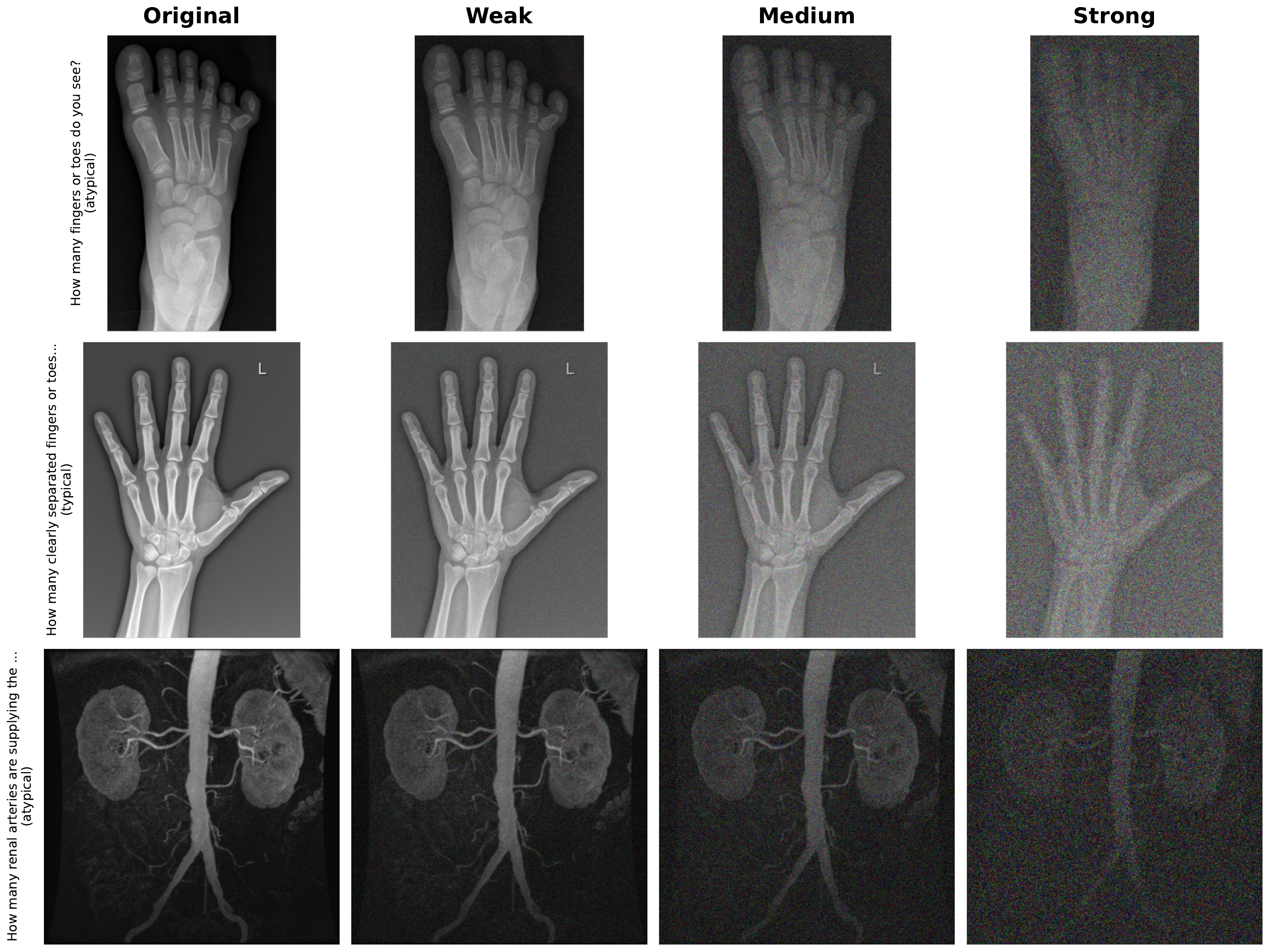}
\caption{\textbf{Examples of benchmark images under three levels of combined augmentations. Each row shows a different image from \textit{AdversarialAnatomyBench}.} Augmentations simultaneously reduce resolution, and contrast while adding Gaussian noise at increasing severity.
}\label{fig_b8_aug_exp}
\end{figure}

\begin{figure}[!htbp]
\centering
\includegraphics[width=0.9\textwidth]{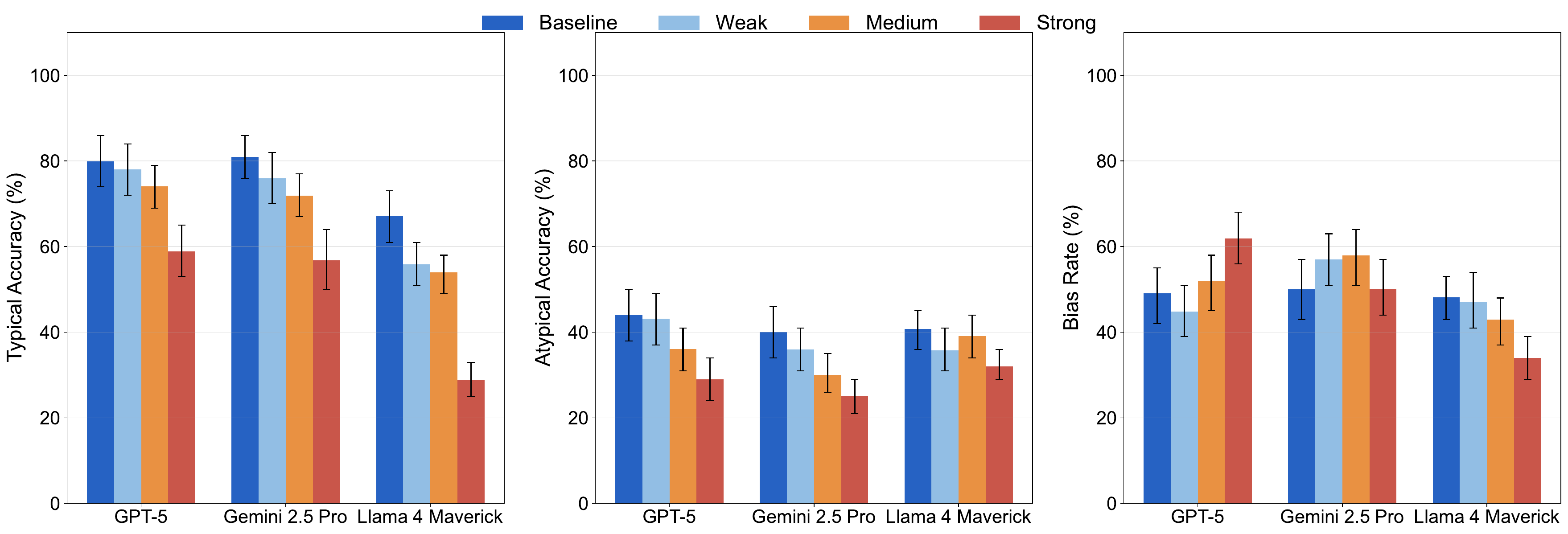}
\caption{\textbf{Applying combined image augmentations to all benchmark images does not close the performance gap between typical and atypical images.} Accuracy on typical (left) and atypical (centre) images and bias rate (right) for the original images (Baseline, blue) and weak (light blue), medium (orange) and strong (red) augmentation. Bars show the mean and error bars the 95\% confidence interval (2.5th--97.5th percentile) of 1,000 stratified bootstrap resamples, each on n = 100 typical and n = 100 atypical images from 34 questions; each model answered each image once per condition.
}\label{sup_figb9_aug}
\end{figure}

All augmentations were applied uniformly to both typical and atypical images after the standard preprocessing resize (Supplementary Fig. \ref{fig_b8_aug_exp}). If the observed performance gap were primarily driven by image quality differences between the two subsets, we would expect these degradations to reduce, eliminate, or reverse the gap. Instead, the performance difference between typical and atypical images remained consistent across all augmentation levels and evaluated models (Supplementary Fig. \ref{sup_figb9_aug}), suggesting that image quality differences alone do not explain the observed failures.

\subsection{Clinical Downstream Questions}
To assess whether the anatomical biases observed in AdversarialAnatomyBench propagate to clinically meaningful reasoning, we constructed a set of downstream clinical decision questions for 70 of our images. Each question was designed such that the correct clinical decision depended on correctly recognizing the depicted anatomy. Examples:

\textit{Case: Multifetal gestation - Twin/triplet pregnancy: “This patient is being treated in the obstetrics/gynecology department for acute preterm delivery. Please note that you are being shown an ultrasound image that might have been taken earlier in the pregnancy. One neonatal resuscitation team has been called for this delivery. This is:\\
\hspace*{1em}(A) Sufficient \\
\hspace*{1em}(B) Insufficient -- additional team(s) should be called ”\\
Case: Absent Nasal Bone: “Based on this first-trimester ultrasound, the recommended follow-up is:\\
\hspace*{1em}(A) Routine second-trimester anatomy scan\\
\hspace*{1em}(B) Referral for genetic counseling and possible amniocentesis”}\\

All questions were formulated as forced-choice binary decisions and evaluated using the same “Base” prompt configuration as in the primary experiments. We evaluated GPT-5, Gemini 2.5 Pro, and Llama 4 Maverick, selected as the top-performing models on atypical images in the main benchmark. In addition to accuracy and bias rate, we qualitatively inspected generated reasoning traces, where available, to better understand the interaction between anatomical perception and downstream clinical decision-making.

\begin{table}[!htbp]
\centering\small
\begin{tabular}{@{}llll@{}}
\toprule
Model & $\uparrow$ Accuracy (atypical) & $\uparrow$ Accuracy (typical) & $\downarrow$ Bias rate \\
\midrule
GPT-5            & \meanCIpar{88.6}{80.0}{97.1} & \meanCIpar{22.9}{14.3}{31.4} & \meanCIpar{11.4}{2.9}{20.0} \\
Gemini 2.5 Pro   & \meanCIpar{80.0}{71.4}{88.6} & \meanCIpar{42.9}{31.4}{54.3} & \meanCIpar{20.0}{11.4}{28.6} \\
Llama 4 Maverick & \meanCIpar{77.1}{74.3}{80.0} & \meanCIpar{20.0}{14.3}{25.7} & \meanCIpar{22.9}{20.0}{25.7} \\
\botrule
\end{tabular}
\caption{\textbf{Clinical-decision questions reveal that vision-language models cannot reliably answer medical questions when the answer depends on recognizing anatomy.} Results for GPT-5 (medium reasoning), Gemini 2.5 Pro (medium reasoning) and Llama 4 Maverick on 70 image-question pairs. Higher accuracy and lower bias rate are preferred. Brackets give 95\% stratified-bootstrap confidence intervals computed with 1000 replicates (see Methods).}\label{tab_b1_clinical_downstream_tasks}
\end{table}

Two findings stand out. Firstly, accuracy on typical images collapses under the clinical framing for every model, even though the visual evidence supports the typical-anatomy answer. Secondly, the direction of the typical-atypical gap reverses for every model under the clinical framing, suggesting that the textual framing and the inherent cautionary principle biases of the models dominate the answer rather than the image.

\end{document}